%% file: arXiv.tex
\documentclass[11pt]{article}
\usepackage[table]{xcolor}
\usepackage{times}
\usepackage[margin=1in]{geometry}
\usepackage{natbib}
\input{math_commands.tex}

\usepackage{url}

\usepackage{booktabs}
\usepackage{multirow}
\usepackage{array}
\usepackage{placeins}
\usepackage{float}
\usepackage{stfloats}
\usepackage{amsmath,amsfonts,amssymb}
\usepackage{textcomp}
\usepackage{verbatim}
\usepackage[caption=false,font=normalsize,labelfont=sf,textfont=sf]{subfig}
\usepackage{adjustbox}
\usepackage{hyperref}
\usepackage{titling}

\pretitle{\begin{center}\LARGE\bfseries}
\posttitle{\par\end{center}\vspace{-0.5em}}
\preauthor{\begin{center}\large}
\postauthor{\par\end{center}\vspace{-1em}}
\predate{}
\postdate{}

\definecolor{bestcell}{RGB}{255,220,220}
\definecolor{secondcell}{RGB}{255,249,200}
\newcommand{\best}[1]{\cellcolor{bestcell}\textbf{#1}}
\newcommand{\second}[1]{\cellcolor{secondcell}\underline{#1}}

\newcommand{\JE}[1]{#1}

\usepackage[normalem]{ulem}
\newif\ifshowchanges
\showchangesfalse
\ifshowchanges
  \DeclareRobustCommand{\deleted}[1]{{\color{black}\sout{#1}}}
  
\else
  \DeclareRobustCommand{\deleted}[1]{}
  
\fi
\title{EvenSplat: Coupled 2D--3D Decomposition
for Gaussian Splatting under Exposure
and Illumination Variation}

\author{
Tongyu Wu$^{1,*}$ \quad
Jacob Edwards$^{2,*}$ \quad
Ziteng Cui$^{3}$ \quad
Caigui Jiang$^{1}$ \quad
Cheng Wang$^{2,\dagger}$\\
$^{1}$Xi'an Jiaotong University \quad
$^{2}$University of East Anglia \quad
$^{3}$University of Tokyo\\
{\footnotesize
$^{*}$Equal contribution \quad
$^{\dagger}$Corresponding author \texttt{(Cheng.C.Wang@uea.ac.uk)}
}
}
\date{}

\begin{document}
\raggedbottom

\maketitle
\begin{abstract}

A surface photographed under even light presents nearly the same appearance from every angle; the same surface under uneven light does not. Exposure changes between views, illumination varies within a single image, and locally strong light sources leave one region bright and its neighbor in shadow. Multi-view reconstruction methods such as 3D Gaussian Splatting treat these lighting artifacts as if they were properties of the scene, entangling capture-specific illumination with the geometry and color they recover. We present EvenSplat, a framework that separates the two. EvenSplat couples an image-space illumination decomposition with an illumination field carried by the Gaussians, so that the same explanation of the lighting is shared between the two-dimensional and three-dimensional views of the scene; a camera-response network and a local exposure-compensation module absorb the global and residual differences that remain across training images. Through extensive experiments across multiple datasets and diverse forms of uneven illumination (cross-view exposure, spatial illumination variation, and high-contrast lighting) on both real-world captured and simulated benchmarks,
EvenSplat generally outperforms state-of-the-art methods, particularly under high-contrast illumination. 

\end{abstract}

\begin{figure}[H]
    \centering
    \includegraphics[width=0.9\textwidth]{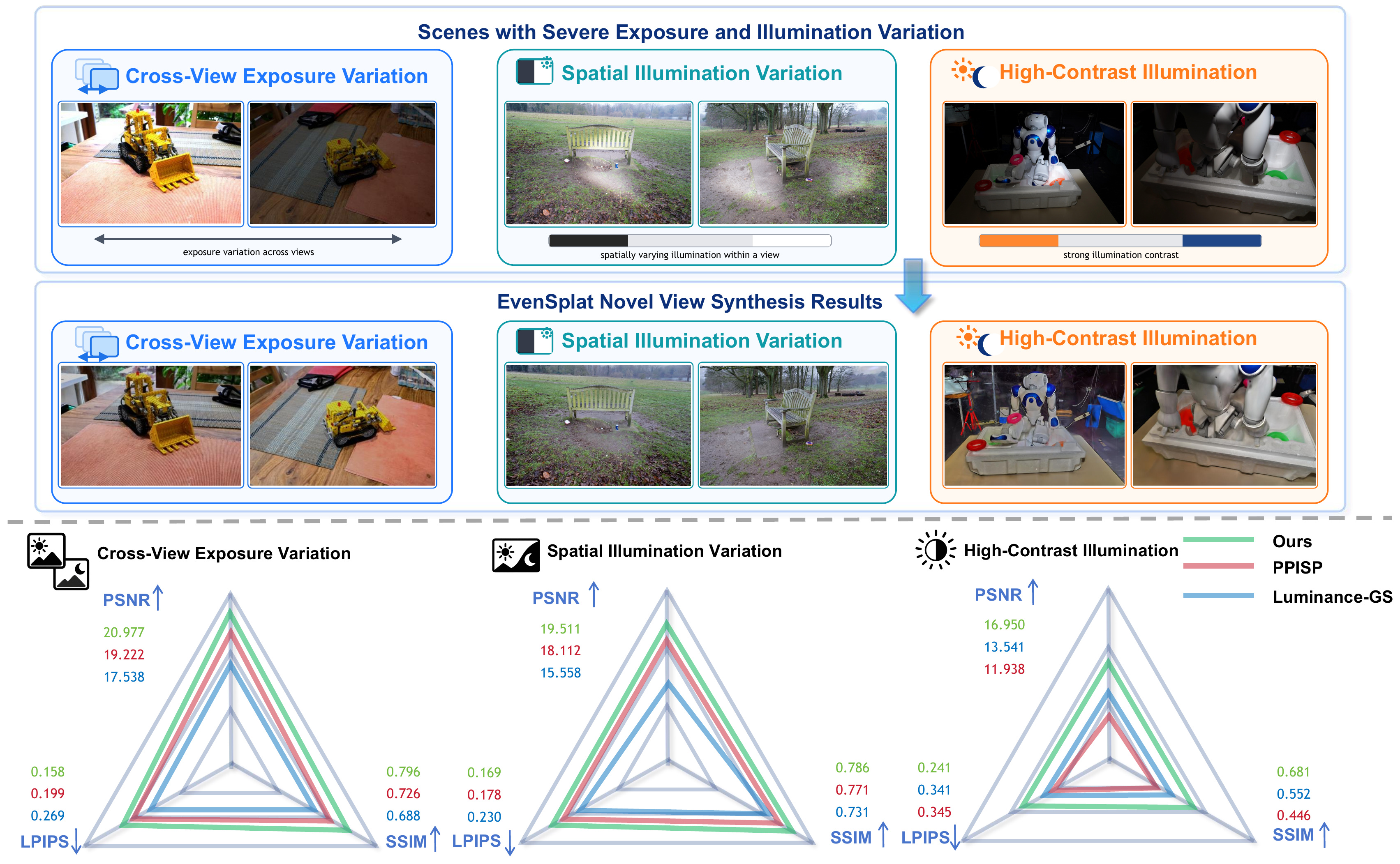}
    \caption{
    EvenSplat reconstructs base appearance from multi-view images with
    cross-view exposure variation (CEV), spatial illumination variation (SIV),
    and high-contrast illumination (HCI).
    Our method generally outperforms the state-of-the-art methods in PSNR, SSIM, and LPIPS on simulated and real-world 3D Gaussian Splatting novel view synthesis.
    }
    \label{fig:teaser}
\end{figure}

\section{Introduction}
\label{sec:intro}

Photographs of the same scene are rarely taken under the same light. A camera's exposure adapts from shot to shot~\citep{shim2014autoexposure}, a flash or a window lights one side of an object more than the other~\citep{debevec1997hdr}, and a bright sky or lamp splits a single frame into washed-out and shadowed regions~\citep{petschnigg2004flash}. For a human these differences are unremarkable; for a system that reconstructs a scene from many such photographs, they are a problem. Each image reports a different brightness and color for the same physical surface, and the reconstruction is left to reconcile observations that no longer agree.

Novel View Synthesis (NVS) methods such as Neural Radiance Fields (NeRF)~\citep{DBLP:journals/corr/abs-2003-08934} and 3D Gaussian Splatting (3DGS)~\citep{kerbl20233dgaussiansplattingrealtime} reconstruct such scenes with high fidelity, but they assume photometric consistency across views. When the inputs disagree, fitting them faithfully entangles the lighting of each capture with the geometry and color of the scene, so the recovered model carries the artifacts of how it was photographed rather than what was photographed.
These inconsistencies take three characteristic forms: \emph{Cross-View Exposure Variation (CEV)}, as when a camera re-meters between shots; \emph{Spatial Illumination Variation
(SIV)} within a single view, as under a flash or a side window; and \emph{High-Contrast
Illumination (HCI)} with strong bright–dark imbalance, as under a harsh local light source. The first is a property of the camera, the latter two of the scene, but all three break the photometric agreement that reconstruction relies on (Fig.~\ref{fig:teaser}).

To evaluate these conditions beyond digital image perturbations, we assembled
a nine-scene real-world dataset with exposure-bracketed and flash-illuminated
captures. It covers the three settings with three CEV, four SIV, and two HCI
scenes, together with evenly illuminated reference captures for appearance evaluation. This dataset
forms a central part of our evaluation, complemented by controlled simulated
benchmarks and a directional-lighting robustness experiment.

We introduce \textbf{EvenSplat}, a coupled 2D--3D appearance decomposition
framework inspired by the reflect- ance--illumination factorization of Retinex
theory~\citep{land1977retinex}. Its central design is to use an
\textit{image-space decomposition} optimized together with the Gaussian
representation. A multi-scale network
predicts per-image illumination and derives a base-appearance estimate, while
a \textit{Gaussian-level illumination field} predicts illumination coefficients
conditioned on position, surface normal, and viewing direction. Aligning the
image-space and rendered illumination maps connects the two branches.
Our contributions are:
(1) A coupled 2D--3D decomposition framework that transfers online
   image-space illumination guidance to a Gaussian-level illumination field
   for reconstruction from exposure- and illumination-inconsistent inputs.
(2) An image-specific correction combining global camera response and illumination-guided local exposure compensation for residual variation in training views.
(3) A nine-scene real-world captured dataset spanning cross-view exposure
   variation, spatial illumination variation, and high-contrast illumination,
   with reference captures for evaluating appearance reconstruction. 

\section{Related Work}
\label{sec:related}

\subsection{NVS in Challenging Lighting Conditions}

Recent ``in-the-wild'' NVS methods explicitly model appearance variations
that violate the photometric consistency assumed by standard NeRF and 3DGS.
NeRF-W~\citep{martinbrualla2020nerfw} introduced per-image appearance
embeddings for unconstrained photo collections. GS-W, SWAG, and WildGaussians
\citep{zhang2024gs-w,dahmani2024swag,kulhanek2024wildgaussians} transfer this
idea to Gaussian scenes, while WildSplat and WilLaGS
\citep{zhang2026wildsplat,bai2026willags} address unposed or strongly
appearance-varying inputs. 

A complementary line targets low-light image formation and exposure.
RawNeRF~\citep{mildenhall2022rawnerf} and HDR-NeRF~\citep{huang2022hdr} model
HDR image formation, while HDR-GS~\citep{hdr_gs} and
PhysHDR-GS~\citep{zeng2026physhdrgs} adapt this objective to Gaussian
representations. Lighting-up NeRF~\citep{wang2023lighting},
Aleth-NeRF~\citep{cui2023aleth}, and Gaussian-DK~\citep{ye2024gaussiandk}
instead target dark observations. LITA-GS~\citep{zhou2025litags} introduces
illumination-invariant physical priors, while LL-Gaussian
\citep{sun2025llgaussian} separates reflectance, illumination, and transient
residuals. Luminance-GS~\citep{cui_luminance_gs},
PPISP~\citep{deutsch2026ppisp}, and P2GS~\citep{shimomura2026p2gs} focus on
exposure, camera processing, and tone-mapping variation.

Other methods address view harmonization or explicitly changing illumination.
CHROMA~\citep{shin2026chroma} harmonizes multi-view inputs using predicted
bilateral grids. DarkGS and SunFaded
\citep{zhang2024darkgs,chang2026sunfaded} model camera-mounted moving lights,
whereas L2DGS and Disco-GS~\citep{kumar2026l2dgs,kumar2026discogs} address
dynamic or colored lighting. 

A related but distinct line performs inverse rendering for relighting. GS-IR,
Relightable 3D Gaussians, and GI-GS
\citep{liang2024gsir,gao2024relightable3dgs,chen2025gigs} recover normals,
materials, visibility, and illumination for relighting. Subsequent methods
extend this factorization to spatially varying materials, indirect illumination,
and unconstrained lights~\citep{sun2025svgir,du2025gsid,zhang2025surgs}.

Even these advances remain brittle in extreme spatially uneven lighting, for instance, scenes with severe cross-view exposure discontinuities, mixed HDR regions, and strong, locally varying light sources across large indoor spaces. In such cases, appearance changes overwhelm per-image embeddings or local feature transfers and drive spurious densification or residual tone inconsistency. 
EvenSplat considers CEV, SIV, and HCI in one
formulation without requiring calibrated light geometry. Instead of aiming to recover a full BRDF or a physically calibrated light, EvenSplat learns an illumination-consistent base appearance for novel-view synthesis by coupling image-space decomposition with
a shared Gaussian illumination field while retaining the efficient, explicit
rendering structure of 3DGS.

\subsection{Image-Space Illumination Decomposition}

Retinex theory \citep{Land1971LightnessAR} separates observed appearance into reflectance and illumination. RetinexNet, KinD, and
URetinex-Net~\citep{wei2018deep,zhang2019kindling,wu2022uretinex} replace
handcrafted decomposition priors with learned illumination estimation for
low-light enhancement. Later work broadens the restoration model:
Retinexformer~\citep{cai2023retinexformer} uses illumination-guided attention;
RetinexMamba and ECMamba~\citep{bai2024retinexmamba,dong2024ecmamba} employ
state-space architectures; and Fourier, video, and diffusion formulations
address mixed exposure and general illumination degradation
\citep{huang2022deepfourier,liu2024vecnet,he2025retidiff}.

Beyond Retinex, intrinsic image decomposition methods separate surface appearance from shading rather than
directly producing an enhanced photograph. Recent work
extends the classical albedo--shading formulation toward more complex
illumination and richer intrinsic properties
\citep{careagaColorful,Dirik_2026_CVPR}.
Related multi-view inverse-rendering work further models material, spatial illumination, and camera response from captured images
\citep{Lin_2025_CVPR}.
These decomposition-based formulations motivate our image-space branch, which
separates illumination from base appearance and uses the resulting
decomposition to guide the learning of a shared 3D Gaussian illumination
representation.

\section{EvenSplat}\label{sec:method}

\begin{figure*}[t]
    \centering
    \includegraphics[width=0.95\linewidth]{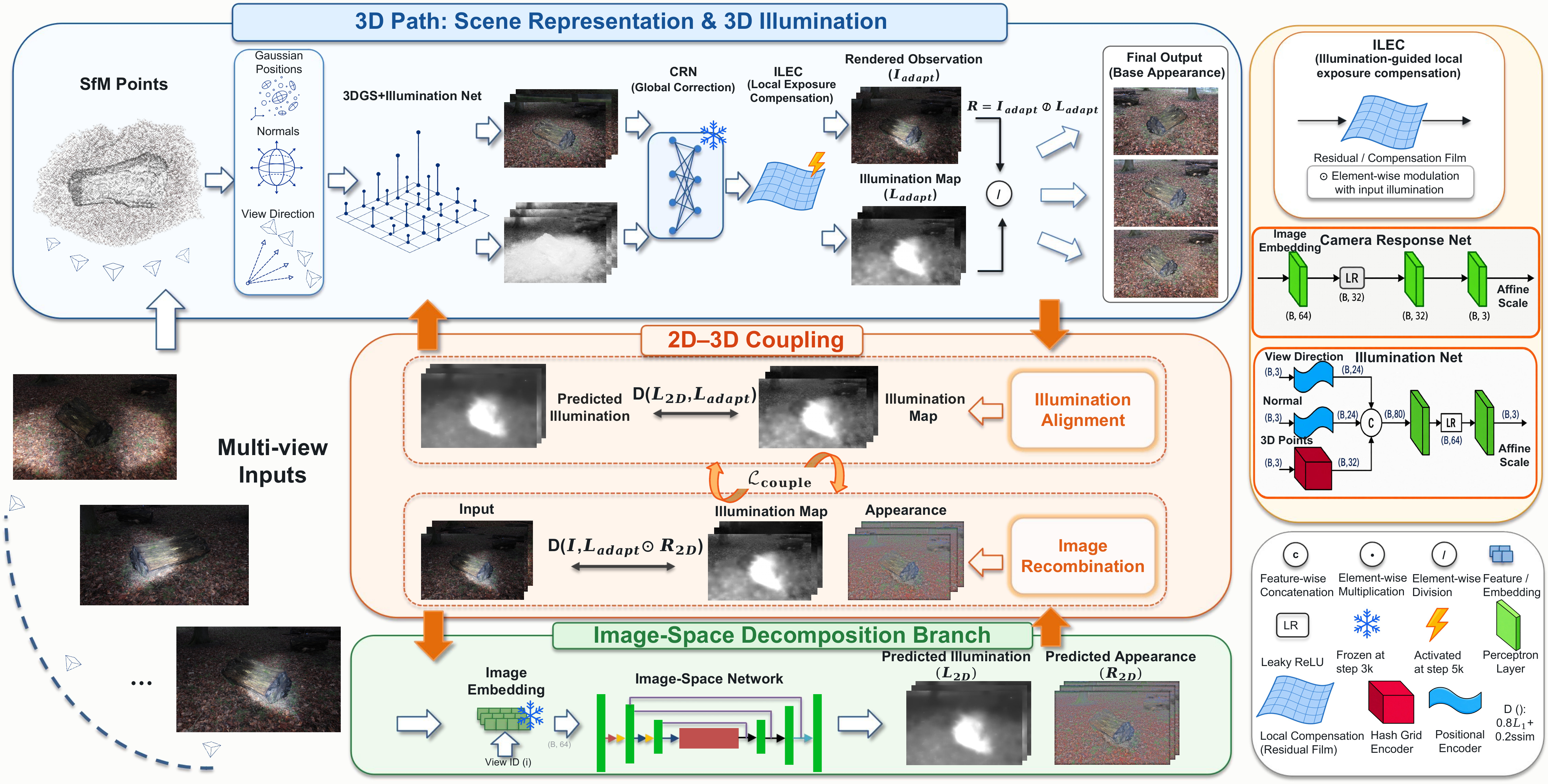}
    \caption{EvenSplat couples image-space decomposition with
    Gaussian-level illumination. Illumination alignment and image recombination connect the branches.
    CRN and ILEC absorb global and spatial training-image residuals. Base appearance is distinguished from the observation-fitting path.
    }
    \label{fig:retinex3d-pipeline}
\end{figure*}

EvenSplat learns a shared Gaussian base-appearance representation from
views with inconsistent exposure and spatially uneven illumination.
 As shown in Fig.~\ref{fig:retinex3d-pipeline},
an online image-space decomposition guides a Gaussian-level illumination field,
while global and local image-specific corrections absorb residual capture
variation during training. The resulting scene representation supports
base-appearance rendering from unseen viewpoints without fitting a test image.

\subsection{Gaussian Appearance and Illumination}
\label{subsec:retinex3d}

Starting from a structure-from-motion point cloud~\citep{schoenberger2016sfm},
we use the standard 3DGS geometry, opacity, and spherical-harmonic color
parameters~\citep{kerbl20233dgaussiansplattingrealtime}. We interpret the SH
color as Gaussian base color, while a shared neural field predicts multiplicative
illumination from Gaussian position, normal, and viewing direction:
\begin{equation}
\mathbf{C}_{j}^{r}=\operatorname{SH}(\mathbf{s}_{j},\mathbf{v}_{j}),
\qquad
\mathbf{a}_{j}=\mathcal{F}_{\mathrm{illum}}
(\boldsymbol{\mu}_{j},\mathbf{n}_{j},\mathbf{v}_{j}).
\label{eq:per_gaussian_illumination}
\end{equation}

Using the same front-to-back compositing weights $w_j^{i}(\mathbf{p})$, we rasterize
the illumination-modulated observation and its illumination map for view $i$:
\begin{equation}
\begin{aligned}
\hat{\mathbf{I}}_{\mathrm{lit}}^{i}(\mathbf{p})
&=\sum_j w_j^{i}(\mathbf{p})
\bigl(\mathbf{a}_{j}\odot\mathbf{C}_{j}^{r}\bigr), 
\qquad
\mathbf{L}_{3D}^{i}(\mathbf{p})
&=\sum_j w_j^{i}(\mathbf{p})\mathbf{a}_{j}.
\end{aligned}
\label{eq:rendered_illumination_map}
\end{equation}

Illumination therefore acts before compositing and is shared across views.
Because the decomposition is learned in sRGB without radiometric calibration and SH base color can remain view-dependent,
we describe $\mathbf{C}_j^r$ as \emph{Gaussian base color}, not physical albedo. Image-specific adaptation (Sec.~\ref{subsec:rlem}) and cross-domain supervision (Sec.~\ref{subsec:coupling}) constrain this ambiguity
without guaranteeing a unique intrinsic decomposition.
Compositing and interpretation details are given in
Appendix~\ref{app:rendering_details}.

\subsection{Image-Specific Appearance Corrections}
\label{subsec:rlem}

The shared illumination field cannot represent every capture-specific
photometric residual. A \emph{camera response network (CRN)} predicts a spatially
uniform per-image gain $\mathbf{c}^{i}$ within the training image-formation path. During training, the CRN compensates for exposure and color-response variations consistently to the rendered
observation and illumination map:
\begin{equation}
\hat{\mathbf{I}}_{\mathrm{cam}}^{i}
=\mathbf{c}^{i}\odot\hat{\mathbf{I}}_{\mathrm{lit}}^{i},
\qquad
\mathbf{L}_{\mathrm{cam}}^{i}
=\mathbf{c}^{i}\odot\mathbf{L}_{3D}^{i}.
\label{eq:camera_response}
\end{equation}

\emph{Illumination-guided local exposure compensation (ILEC)} models remaining
spatial residuals with a low-resolution log-gain map
$\Delta_{\mathrm{loc}}^{i}\in\mathbb{R}^{h\times w\times1}$. Bilinear
upsampling and a gradual activation ramp $r(t)$ yield the adapted observation and illumination map:
\begin{equation}
\begin{aligned}
\hat{\mathbf{I}}_{\mathrm{adapt}}^{i}
=\hat{\mathbf{I}}_{\mathrm{cam}}^{i}\odot
\exp\!\left(r(t)\,\mathrm{Up}(\Delta_{\mathrm{loc}}^{i})\right),\qquad
\mathbf{L}_{\mathrm{adapt}}^{i}
=\mathbf{L}_{\mathrm{cam}}^{i}\odot
\exp\!\left(r(t)\,\mathrm{Up}(\Delta_{\mathrm{loc}}^{i})\right).
\end{aligned}
\label{eq:adapted_illumination}
\end{equation}

The low-resolution parameterization limits texture memorization, while
illumination-guided magnitude, shrinkage, and promotion terms regularize the
map (Appendix~\ref{app:local_losses}). CRN and ILEC absorb image-specific photometric variation during training rather
than becoming part of the transferable scene representation. They affect training but cancel
from the base-appearance ratio because they multiply both paths. At novel views,
their image-specific gains are set to the identity.

\subsection{Online Image-Space Decomposition}
\label{subsec:retinex2d_mechanism}

Reconstruction alone leaves the allocation of variation between predicted base appearance
and illumination ambiguous. The image-space branch guides this separation through online updates during scene fitting. For each training image, a multi-scale image-space network conditioned on a learnable image embedding ($\mathbf{e}^{i}\in\mathbb{R}^{d_{\mathrm{emb}}}$) predicts channel-shared illumination and derives predicted base appearance by division:
\begin{equation}
\mathbf{L}_{2D}^{i}
=0.05+3.95\sigma\!\left(\mathcal{F}_{2D}(\mathbf{I}^{i},\mathbf{e}^{i})\right),
\qquad
\mathbf{R}_{2D}^{i}
=\mathbf{I}^{i}\oslash(\mathbf{L}_{2D}^{i}+\epsilon),
\label{eq:retinex2d_reflectance}
\end{equation}
where $\sigma(x)=1/(1+e^{-x})$ denotes the sigmoid activation function.
The affine scaling constrains $\mathbf{L}_{2D}^{i}$ to $(0.05,4.0)$,
enabling both attenuation and amplification while preventing unstable
near-zero division. The learnable image embedding captures image-specific
exposure and appearance variations through FiLM modulation~\citep{perez2017filmvisualreasoninggeneral}.
The network architecture is in Fig.~9.

\begin{figure}[t]
    \centering
   \includegraphics[width=\linewidth]{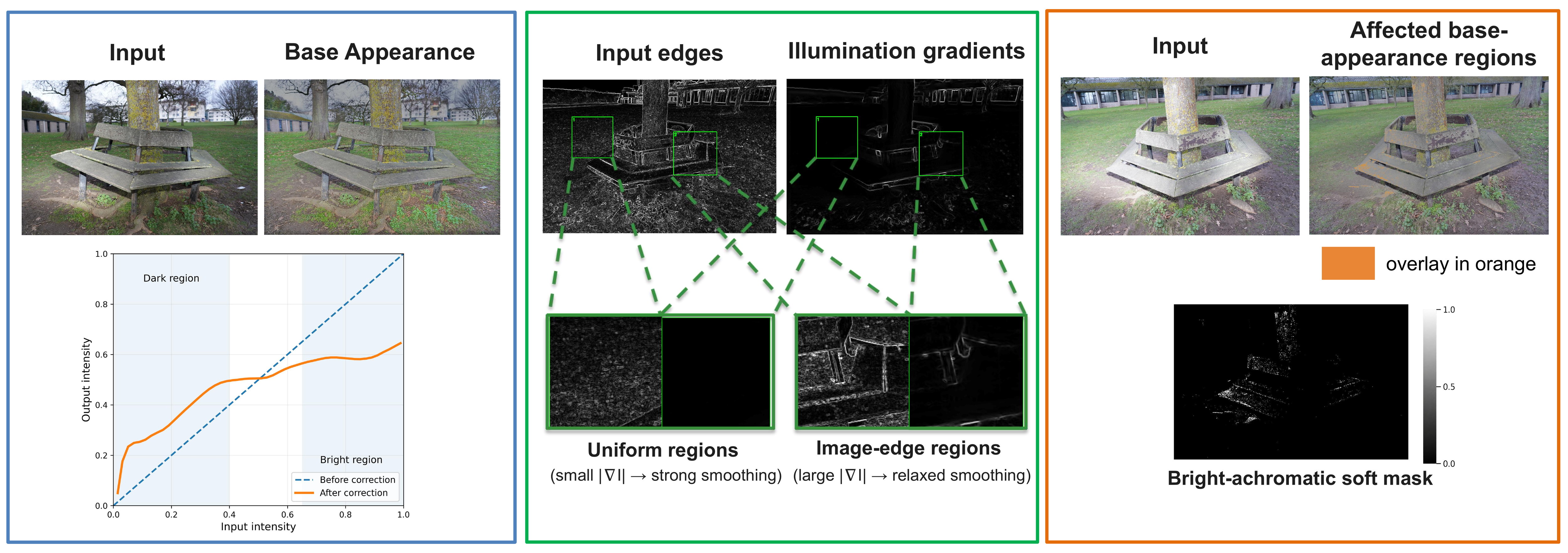}
    \caption{
    Image-space regularization, from left to right: adaptive curve
    constraints on base-appearance intensity; edge-aware illumination
    smoothness; and white preservation using a bright-achromatic soft mask.
    }
    \label{fig:decomposition_priors}
\end{figure}

As illustrated in Fig.~\ref{fig:decomposition_priors}, we regularize the decomposition with three priors. \emph{Adaptive curve regularization}
selects excessively dark or bright base-appearance pixels and pulls them toward
moderate target intensities. \emph{Edge-aware smoothness} penalizes illumination
gradients with weights derived from input-image gradients, encouraging smooth
lighting without blurring observed boundaries. \emph{White preservation}  uses a soft
bright, low-chroma CIELAB mask to retain approximately achromatic regions of
the input. The complete loss definitions and
hyperparameters are given in Appendix~\ref{app:image_losses}. This branch is
optimized jointly with the scene rather than used as independent preprocessing.

\subsection{Coupled Optimization and Rendering}
\label{subsec:coupling}

Our \emph{coupled optimization} connects the image-space decomposition to the
Gaussian illumination representation. We use a common image discrepancy combining $\ell_1$ reconstruction and SSIM:
\begin{equation}
D(\mathbf{A},\mathbf{B})
=(1-\lambda_{\mathrm{ssim}})\|\mathbf{A}-\mathbf{B}\|_1
+\lambda_{\mathrm{ssim}}[1-\operatorname{SSIM}(\mathbf{A},\mathbf{B})].
\label{eq:image_discrepancy}
\end{equation}

The cross-domain \emph{coupled loss} aligns the adapted 3D illumination with the image-space estimate and recombines it with
$\mathbf{R}_{2D}$ to reconstruct the input: 
\begin{equation}
\begin{aligned}
\mathcal{L}_{\mathrm{couple}}=\mathcal{L}_{\mathrm{align}}+\mathcal{L}_{\mathrm{recomb}}
={}&\lambda_{\mathrm{align}}
D(\mathbf{L}_{\mathrm{adapt}},\mathbf{L}_{2D})+\lambda_{\mathrm{recomb}}
D(\mathbf{L}_{\mathrm{adapt}}\odot\mathbf{R}_{2D},\mathbf{I}).
\end{aligned}
\label{eq:coupling_objective}
\end{equation}
Direct alignment compares illumination maps, whereas recombination measures
compatibility in the observed-image domain. 
Both branches enter this
objective and parameters are updated during training when activated.
When activated, no stop-gradient is applied, so gradients from the coupling loss propagate to both the image-space network and the Gaussian illumination branch.

\emph{Observation fidelity} supervises the corrected Gaussian rendering to agree with captured images, $\mathcal{L}_{\mathrm{obs}}
=D(\hat{\mathbf{I}}_{\mathrm{adapt}},\mathbf{I})$. This route retains the captured image as the reconstruction target throughout
scene fitting.

Collecting weighted regularizers into $\mathcal{R}$, we write the objective as\JE{:}
\begin{equation}
\mathcal{L}
=\lambda_{\mathrm{cpl}}\mathcal{L}_{\mathrm{couple}}
+\lambda_{\mathrm{obs}}\mathcal{L}_{\mathrm{obs}}
+\mathcal{R}.
\label{eq:total_objective}
\end{equation}
$\mathcal{R}$ contains the image-space regularizers, Gaussian
appearance--illumination gradient exclusion, higher-order SH regularization,
and ILEC local-map penalties detailed in
Appendix~\ref{app:loss_definitions}. 

\paragraph{Staged fitting.}
The reported training protocol runs for 10,000 iterations. The illumination
field, image embeddings, and CRN are frozen after 3,000 iterations;
ILEC is activated at iteration 5,000. The image-space decomposition network and $\mathcal{L}_{\mathrm{couple}}$ remain active throughout training. Novel-view rendering uses only the
shared Gaussian appearance and illumination field.

\section{Experiments}
\label{sec:experiments}

We test the central claim behind EvenSplat: coupling image-space priors with a
shared 3D illumination model is most useful when photometric variation cannot
be explained by a global per-view correction. We evaluate physically captured
CEV, SIV, and HCI scenes, repeat the comparison under controlled perturbations,
and use cross-lighting and component ablations to probe what is learned.

\subsection{Datasets}
\label{subsec:datasets}

\begin{figure}[h]
    \centering
    \includegraphics[width=0.98\linewidth]{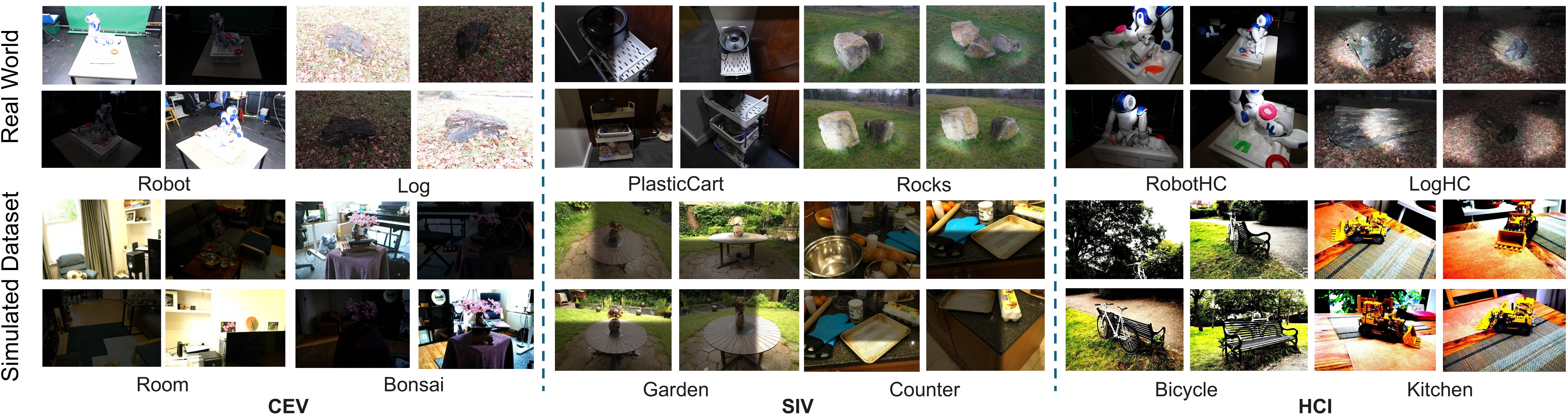}
    \smallskip
    \caption{Gallery of the three illumination settings in the real-world and simulated datasets. CEV changes
    exposure across views, SIV introduces spatial illumination variation within
    images, and HCI produces a strong bright--dark imbalance. 
    }
    \label{fig:datasets}
\end{figure}

\paragraph{Real-world benchmark.}
Our nine-scene benchmark contains three Cross-View Exposure Variation (CEV),
four Spatial Illumination Variation (SIV), and two High-Contrast Illumination
(HCI) scenes spanning indoor and outdoor subjects. Unlike a purely synthetic
benchmark, these captures retain camera response, highlight saturation, flash
falloff, and cast shadows. A second collection under more uniform illumination
provides an appearance reference for each scene.
Figure~\ref{fig:datasets} demonstrates the three settings; capture parameters,
protocols, and structure-from-motion initialization are reported in
Appendix~\ref{app:capture_real} and Table~\ref{tab:capture-details}.

\paragraph{Simulated benchmark.}
We additionally perturb six Mip-NeRF~360 scenes~\citep{barron2022mipnerf360}
while retaining the original images as references. CEV follows the varying-exposure benchmark
of~\citet{cui_luminance_gs}; SIV applies smoothly blended exposure factors to
two image regions; and HCI increases training-image contrast. Exact operators are in Appendix~\ref{app:capture_simulated}.

\subsection{Experimental Results}
\label{subsec:experimental_results}

\paragraph{Training and evaluation protocol.}
All training and evaluation runs are conducted on two NVIDIA GeForce
RTX~3090 GPUs. EvenSplat is implemented with gsplat~\citep{ye2025gsplat} and
trained for 10,000 iterations using the staged fitting in Sec.~\ref{subsec:coupling}.
We compare against 3DGS~\citep{kerbl20233dgaussiansplattingrealtime}, 3DGS+CHROMA~\citep{shin2026chroma}, GS-W~\citep{zhang2024gs-w}, Bilateral Grid
~\citep{wang2024bilateral}, PPISP~\citep{deutsch2026ppisp}, and Luminance-GS~\citep{cui_luminance_gs}. Full settings
are in Appendix~\ref{app:implementation}.

\begin{figure}[thb]
    \centering
    \includegraphics[width=\linewidth]{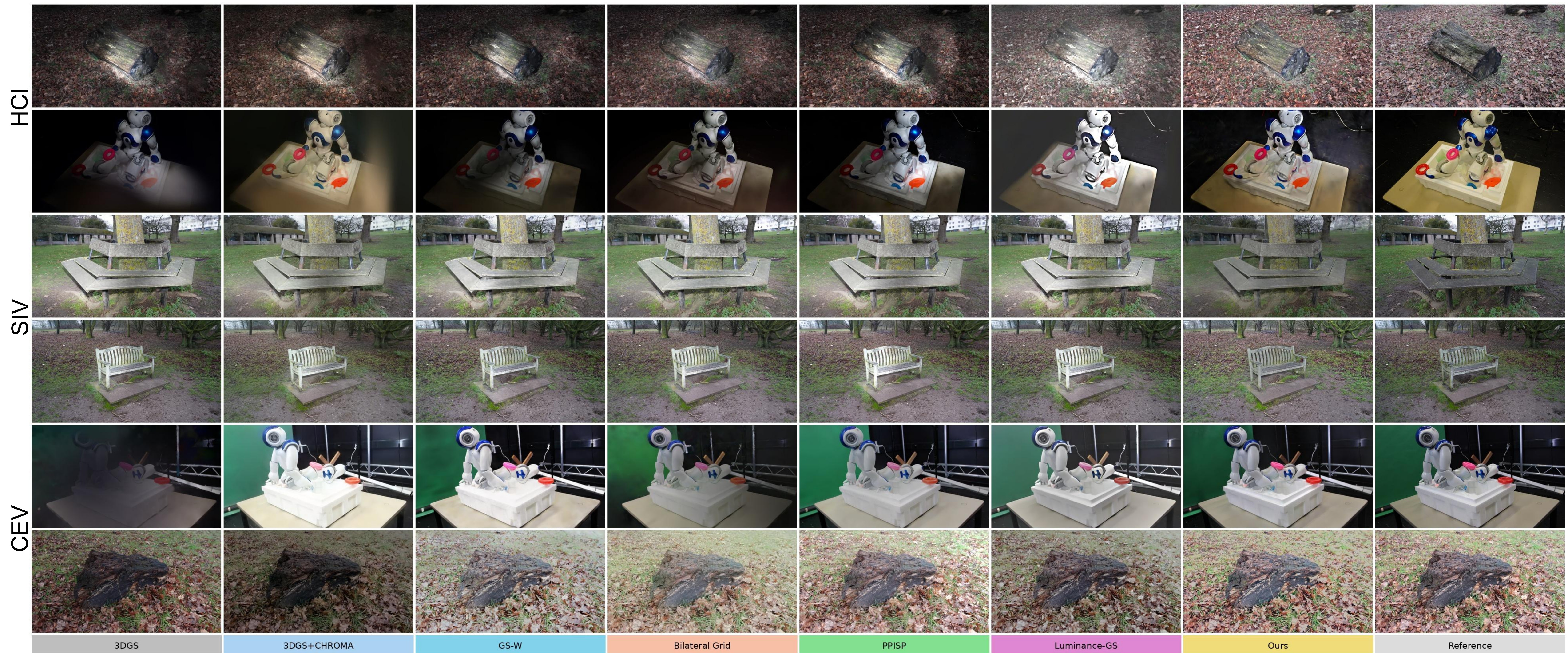}
    \caption{Real-world comparisons. Columns show 3DGS, 3DGS+CHROMA, GS-W, Bilateral Grid,
    PPISP, Luminance-GS, EvenSplat, and the captured reference. Rows 1--2 show
    HCI, rows 3--4 SIV, and rows 5--6 CEV. EvenSplat most clearly separates
    appearance from illumination in the spatially uneven HCI and SIV.}
    \label{fig:comp-real}
\end{figure}

\paragraph{Real-world reconstruction.}
Tables~\ref{tab:average-comparison} and~\ref{tab:real-world-results} reveal a
clear condition-dependent pattern. EvenSplat leads every metric for SIV and
HCI, including a 3.17~dB PSNR gain over the strongest HCI baseline and a
1.40~dB gain over PPISP for SIV. These are precisely the regimes in which a
global exposure or ISP correction cannot explain illumination that changes
within an image. Figure~\ref{fig:comp-real} shows the corresponding qualitative distinction:
several baselines either preserve the capture shadow or flatten bright regions,
whereas EvenSplat recovers detail across both sides of the illumination
boundary. For the simpler CEV setting, EvenSplat obtains the best PSNR
but PPISP retains better SSIM and LPIPS. 

To isolate whether the improvements
come merely from the image-space tone prior, Appendix
Table~\ref{tab:all-results} additionally reports \emph{Tone-Matched 3DGS},
which applies a single global tone/exposure normalization to vanilla 3DGS renderings using the same prescribed mid-tone target of 0.5.
This control improves PSNR on two of the three CEV
scenes, but it underperforms vanilla 3DGS on every SIV and HCI scene and remains
well below EvenSplat. The gains in these more challenging settings therefore
cannot be explained by global tone matching alone.

\begin{table*}[t]
\centering
\begin{minipage}[t]{0.49\textwidth}
\centering
\caption{Scene-averaged real-world results.
Entries are PSNR$\uparrow$/SSIM$\uparrow$/LPIPS$\downarrow$.
Best values are bold with light-red shading; second-best values are
underlined with light-yellow shading.}
\label{tab:average-comparison}
\begingroup
\fontsize{6}{7}\selectfont
\setlength{\tabcolsep}{0pt}
\renewcommand{\arraystretch}{1.08}
\begin{adjustbox}{max width=\linewidth}
\begin{tabular}{@{}l@{\hspace{3pt}}r@{/}r@{/}r@{\hspace{4pt}}r@{/}r@{/}r@{\hspace{4pt}}r@{/}r@{/}r@{}}
\toprule
\textbf{Method} & \multicolumn{3}{c}{\shortstack{\textbf{CEV}\\(3 scenes)}} & \multicolumn{3}{c}{\shortstack{\textbf{SIV}\\(4 scenes)}} & \multicolumn{3}{c}{\shortstack{\textbf{HCI}\\(2 scenes)}} \\
\midrule
3DGS
& 14.197 & 0.461 & 0.609
& 17.902 & 0.734 & 0.250
& 11.702 & 0.427 & 0.390 \\

3DGS+CHROMA
& 10.193 & 0.530 & 0.415
& 16.932 & 0.657 & 0.239
& \second{13.780} & 0.522 & 0.352 \\

Bilateral Grid
& 16.337 & 0.549 & 0.524
& 17.994 & 0.743 & 0.227
& 12.190 & 0.404 & 0.349 \\

GS-W
& 12.404 & 0.523 & 0.362
& 17.742 & 0.702 & 0.243
& 11.801 & 0.425 & 0.379 \\

Luminance-GS
& 16.585 & 0.538 & 0.485
& 15.558 & 0.731 & 0.230
& 13.541 & \second{0.553} & \second{0.341} \\

PPISP
& \second{17.794} & \best{0.723} & \best{0.269}
& \second{18.112} & \second{0.771} & \second{0.178}
& 11.938 & 0.446 & 0.345 \\

EvenSplat
& \best{19.526} & \second{0.708} & \second{0.305}
& \best{19.511} & \best{0.786} & \best{0.169}
& \best{16.950} & \best{0.682} & \best{0.241} \\
\bottomrule
\end{tabular}
\end{adjustbox}
\endgroup
\end{minipage}
\hfill
\begin{minipage}[t]{0.49\textwidth}
\centering
\caption{Scene-averaged results on simulated data.
Entries are PSNR$\uparrow$/SSIM$\uparrow$/LPIPS$\downarrow$.
Best values are bold with light-red shading; second-best values are
underlined with light-yellow shading.}
\label{tab:simulated-comparison}
\begingroup
\fontsize{6}{7}\selectfont
\setlength{\tabcolsep}{0pt}
\renewcommand{\arraystretch}{1.08}
\begin{adjustbox}{max width=\linewidth}
\begin{tabular}{@{}l@{\hspace{3pt}}r@{/}r@{/}r@{\hspace{4pt}}r@{/}r@{/}r@{\hspace{4pt}}r@{/}r@{/}r@{}}
\toprule
\textbf{Method} & \multicolumn{3}{c}{\shortstack{\textbf{CEV}\\(6 scenes)}} & \multicolumn{3}{c}{\shortstack{\textbf{SIV}\\(6 scenes)}} & \multicolumn{3}{c}{\shortstack{\textbf{HCI}\\(6 scenes)}} \\
\midrule
3DGS  & 17.288 & 0.605 & 0.312 & 18.238 & 0.773 & 0.193 & 13.000 & 0.452 & 0.358 \\
3DGS+CHROMA & 17.566 & \second{0.750} & \second{0.198} & 18.500 & 0.745 & 0.240 & 14.197 & 0.502 & 0.341 \\
Bilateral Grid &  18.777 & 0.674 & 0.253 & 19.102 & 0.773 & \second{0.144} & 14.056 & 0.496 & 0.322 \\
GS-W  & 17.517 & 0.640 & 0.215 & 18.371 & 0.776 & 0.173 & 12.987 & 0.451 & 0.360 \\
Luminance-GS  & 17.538 & 0.688 & 0.269 & 14.485 & 0.682 & 0.229 & 12.963 & \second{0.568} & 0.347 \\
PPISP  & \second{19.222} & 0.726 & 0.199 & \second{20.343} & \second{0.814} & 0.151 & \second{15.174} & \best{0.589} & \second{0.302} \\
EvenSplat & \best{20.977} & \best{0.796} & \best{0.158} & \best{22.000} & \best{0.862} & \best{0.097} & \best{16.736} & 0.501 & \best{0.301} \\
\bottomrule
\end{tabular}
\end{adjustbox}
\endgroup
\end{minipage}
\end{table*}

\begin{table*}[tbh]
\centering
\caption{Selected real-world scenes under CEV, SIV, and HCI.
Each method entry reports PSNR$\uparrow$/SSIM$\uparrow$/LPIPS$\downarrow$.
Best values are bold with light-red shading; second-best values are underlined
with light-yellow shading. Ties at the reported precision share the same rank.
All nine scenes and the complete evaluation
results are reported in Appendix Table~\ref{tab:all-results}.}
\label{tab:real-world-results}
\begingroup
\fontsize{7}{8}\selectfont
\setlength{\tabcolsep}{3pt}
\setlength{\fboxsep}{0.7pt}
\renewcommand{\arraystretch}{1.16}
\newcommand{\bestnum}[1]{\colorbox{bestcell}{\textbf{#1}}}
\newcommand{\secondnum}[1]{\colorbox{secondcell}{\underline{#1}}}
\begin{adjustbox}{width=\textwidth}
\begin{tabular}{@{}llcccccccc@{}}
\toprule
\textbf{Setting} & \textbf{Scene} & \textbf{3DGS} & \textbf{3DGS+CHROMA} & \textbf{Bilateral Grid}
& \textbf{GS-W} & \textbf{Luminance-GS} & \textbf{PPISP} & \textbf{EvenSplat} \\
\midrule

\multirow{2}{*}{\textbf{HCI}}
& \textit{LogHC}
& 10.533/0.393/0.328
& 10.597/0.380/0.350
& 11.312/0.438/0.302
& 10.686/0.396/0.325
& \secondnum{12.614}/\secondnum{0.565}/\secondnum{0.287}
& 10.568/0.400/0.322
& \bestnum{16.718}/\bestnum{0.669}/\bestnum{0.220} \\

& \textit{RobotHC}
& 12.870/0.460/0.452
& 16.963/0.664/0.353
& 13.068/0.370/0.395
& 12.916/0.454/0.432
& \secondnum{14.468}/\secondnum{0.540}/0.394
& 13.307/0.491/\secondnum{0.367}
& \bestnum{17.181}/\bestnum{0.694}/\bestnum{0.262} \\

\midrule

\multirow{2}{*}{\textbf{SIV}}
& \textit{CircularBench}
& 15.953/0.702/0.244
& 15.056/0.661/0.275
& \secondnum{16.147}/0.690/0.233
& 16.134/0.654/0.257
& 14.928/0.725/0.214
& 15.854/\bestnum{0.741}/\secondnum{0.191}
& \bestnum{17.420}/\secondnum{0.739}/\bestnum{0.186} \\

& \textit{ParkBench}
& 20.310/\secondnum{0.810}/\secondnum{0.135}
& \secondnum{20.828}/0.785/0.165
& 19.907/0.803/0.138
& 20.043/0.761/0.168
& 18.248/0.803/0.136
& 19.810/\bestnum{0.816}/0.142
& \bestnum{21.037}/\bestnum{0.816}/\bestnum{0.117} \\

\midrule

\multirow{2}{*}{\textbf{CEV}}
& \textit{Robot}
& 14.564/0.601/0.519
& 13.060/0.666/0.271
& 16.542/0.677/0.429
& 14.814/0.798/0.224
& 16.325/0.656/0.304
& \secondnum{21.829}/\secondnum{0.861}/\secondnum{0.133}
& \bestnum{22.781}/\bestnum{0.884}/\bestnum{0.097} \\

& \textit{Log}
& 17.072/0.651/0.367
& 10.503/0.547/0.294
& 20.130/0.783/0.216
& 13.432/0.676/0.220
& 17.745/0.734/0.283
& \secondnum{20.547}/\bestnum{0.871}/\bestnum{0.115}
& \bestnum{20.648}/\secondnum{0.812}/\secondnum{0.157} \\

\bottomrule
\end{tabular}
\end{adjustbox}
\endgroup
\end{table*}

\paragraph{Controlled simulated reconstruction.}
The controlled results in Table~\ref{tab:simulated-comparison} show that the
real-world gains are not specific to our capture pipeline. EvenSplat leads all
metrics for CEV and SIV, improving PSNR over PPISP by 1.76 and 1.66~dB,
respectively. Under HCI it retains the best PSNR (+1.56~dB) but trails PPISP in
SSIM, indicating that correction under severe clipping trades pixel
agreement against structural similarity. Per-scene results in
Appendix~\ref{app:simulated_results} show that this trend is not driven by a single scene.

\begin{figure}[t]
    \centering
    \includegraphics[width=\linewidth]{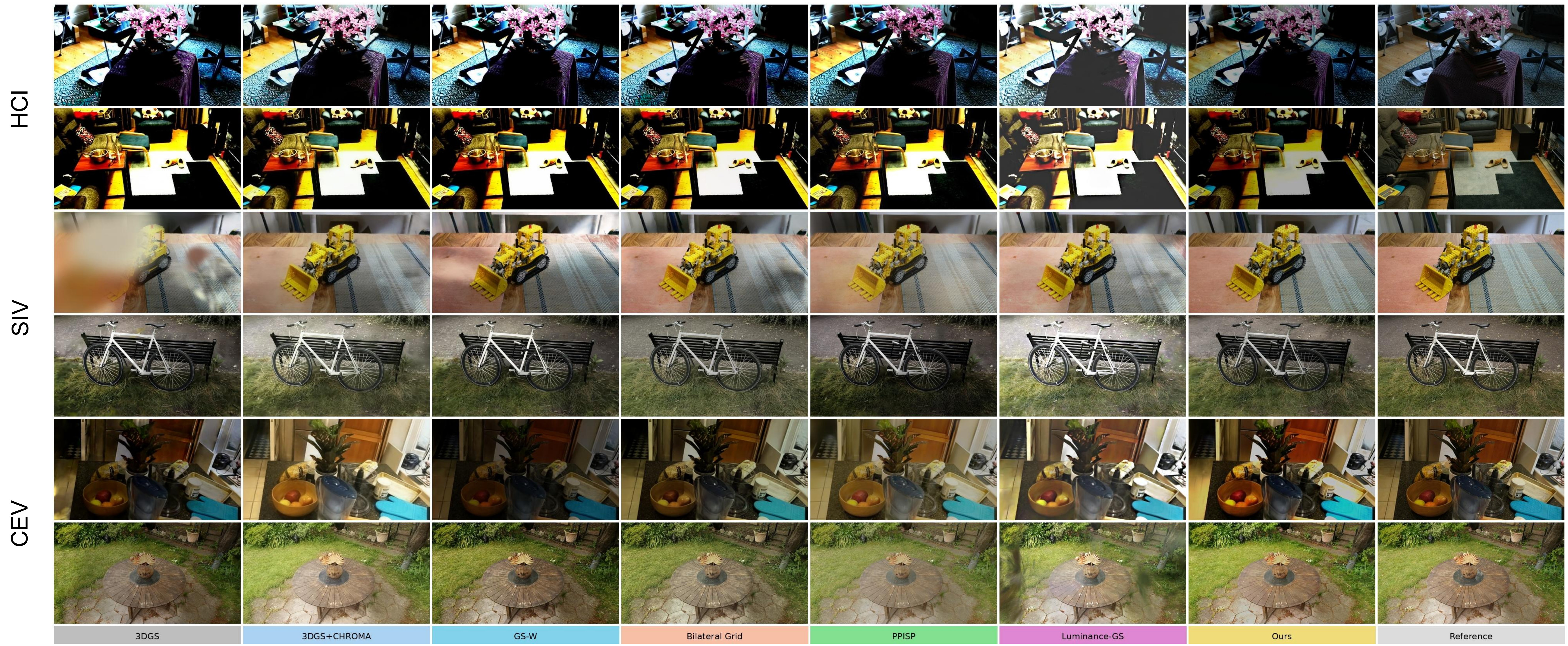}
    \caption{Simulated HCI, SIV, and CEV comparisons. EvenSplat reduces
    illumination leakage while retaining scene texture; HCI remains the most
    challenging setting because contrast amplification clips information.}
    \label{fig:comp-simulated-selected}
\end{figure}

\paragraph{Cross-lighting appearance consistency.}
This test removes the need to choose a lighting-neutral
reference: The two capture sets differ only in illumination
direction, with light incident predominantly from the left or right, while
scene geometry and materials remain unchanged. We therefore assess whether
reconstructions from these differently illuminated observations yield similar
base appearance. 

\begin{figure}[tbh]
    \centering
    \includegraphics[width=\linewidth]{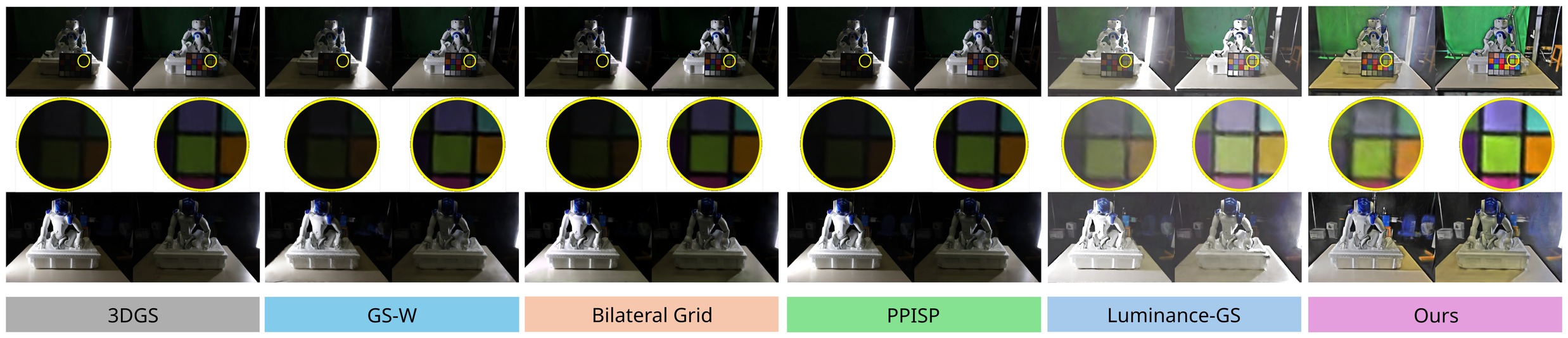}
    \caption{Cross-lighting appearance consistency. The capture sets
    differ only in left- versus right-dominant illumination. Paired appearance
    renderings and color-chart crops show the resulting reconstructions.
    PSNR is computed between corresponding left- and right-dataset appearance
    renderings, using one as the reference for the other; higher values
    indicate greater consistency across lighting conditions. PSNR (dB):
    3DGS 10.5628; GS-W 10.6427;
    Bilateral Grid 10.3909; PPISP 10.3641; Luminance-GS
    \underline{12.7311} (second-best); EvenSplat \textbf{15.5723} (best).
    }
    \label{fig:directional-lighting}
\end{figure}

EvenSplat reaches 15.57~dB, 2.84~dB
above Luminance-GS, the strongest baseline.
The gain indicates that the recovered representation is less tied to the
direction of the capture light. The score alone
cannot establish the accuracy of the recovered base appearance, Figure~\ref{fig:directional-lighting} compares the resulting
appearance renderings and color-chart crops. 

\paragraph{Dual outputs evaluation on HDR-NeRF dataset.} Our method design makes prediction on both \emph{Rendered Observation} and \emph{Base Appearance} possible. On the simulated HDR-NeRF dataset~\citep{huang2022hdr}, illumination is severe
but consistent across views. EvenSplat retains high-quality full-scene
rendering (37.6 dB mean PSNR across scenes) while additionally extracting base appearance,
with the highest base appearance PSNR (Appendix Table~\ref{tab:hdr-base-appearance}) on every scene against ground truth under
parallel uniform light. If the shared light saturates every view, clipping removes the original
color for all methods---an identifiability limit of the observations, not a
method-specific failure. Full results are in Appendix~\ref{app:simulated_results}.

\subsection{Ablation Study}
\label{subsec:ablations}

\paragraph{Coupled model and Gaussian-level components.}
Table~\ref{tab:retinex3d-ablation} supports this design. Removing image-space
decomposition costs 3.60~dB on CEV and 4.94~dB on HCI, yet using that branch
alone is also substantially worse: the 2D prior needs a shared 3D explanation.
CRN removal degrades every setting, confirming the role of global per-image
correction. ILEC is sharply specialized: removing it costs 2.88~dB on HCI but
is nearly neutral on CEV and SIV. Thus global response, shared illumination,
and local compensation address different error scales; the full model offers
the most reliable balance across conditions. Appendix~\ref{app:simulated_results}
further isolates the ILEC regularizers.

\begin{table*}[h]
\centering
\begin{minipage}[t]{0.49\textwidth}
\caption{Ablation of EvenSplat on the real-world datasets (PSNR/SSIM/LPIPS).}
\label{tab:retinex3d-ablation}
\centering
\setlength{\tabcolsep}{2pt}
\renewcommand{\arraystretch}{1.08}
\begin{adjustbox}{max width=\linewidth}
\fontsize{6}{7}\selectfont
\begin{tabular}{lccc}
\hline\hline
\textbf{Variant}
& \textbf{CEV}
& \textbf{SIV}
& \textbf{HCI} \\
\hline

EvenSplat (Full)
& 19.526/0.708/0.305
& 19.511/0.786/0.169
& 16.950/0.682/0.241 \\

No Image-Space Decomp.
& 15.924/0.646/0.374
& 19.216/0.710/0.242
& 12.012/0.438/0.324 \\

No Normals
& 16.839/0.661/0.383
& 19.506/0.782/0.172
& 16.712/0.674/0.246 \\

No View Directions
& 16.916/0.665/0.366
& 19.601/0.786/0.165
& 16.745/0.670/0.247 \\

No CRN
& 16.202/0.649/0.411
& 18.094/0.766/0.185
& 15.184/0.632/0.280 \\

No ILEC
& 19.599/0.707/0.308
& 19.467/0.782/0.168
& 14.066/0.561/0.266 \\

Image-Space Decomp. Only
& 15.632/0.587/0.382
& 12.601/0.483/0.583
& 15.092/0.531/0.425 \\

\hline\hline
\end{tabular}
\end{adjustbox}
\end{minipage}
\hfill
\begin{minipage}[t]{0.49\textwidth}
\caption{Ablation of the image-space decomposition branch on the simulated dataset.
}
\label{tab:retinex2d-ablation}
\centering
\setlength{\tabcolsep}{2pt}
\renewcommand{\arraystretch}{1.08}
\begin{adjustbox}{max width=\linewidth}
\fontsize{6}{7}\selectfont
\begin{tabular}{lccc}
\hline\hline
\textbf{Variant} & \textbf{CEV} & \textbf{SIV} & \textbf{HCI} \\
\hline
Image-Space Decomp. Only
& 17.962/0.706/0.240
& 15.260/0.676/0.263
& 16.478/0.547/0.376 \\

\shortstack[l]{No Loss\_\\(Adaptive Curve)}
& 17.786/0.702/0.249
& 14.213/0.622/0.284
& 13.540/0.524/0.396 \\

\shortstack[l]{No Loss\_\\(Smooth Edge-Aware)}
& 17.568/0.560/0.405
& 13.922/0.480/0.452
& 16.508/0.504/0.434 \\

\shortstack[l]{No Loss\_\\(White Preservation)}
& 17.118/0.688/0.237
& 14.245/0.626/0.284
& 15.063/0.545/0.380 \\

Colored Illumination
& 17.809/0.673/0.239
& 15.420/0.671/0.229
& 17.086/0.526/0.308 \\

\hline\hline
\end{tabular}
\end{adjustbox}
\end{minipage}
\end{table*}
\paragraph{Image-space decomposition components.}
Table~\ref{tab:retinex2d-ablation} separates the roles of the three priors in
Fig.~\ref{fig:decomposition_priors}. Adaptive curve regularization is most
important under HCI, where its removal costs 2.94~dB. Edge-aware smoothness
drives structural and perceptual quality, producing the largest SSIM and LPIPS
degradation under CEV and SIV when removed. White preservation provides a
smaller but consistent PSNR benefit. Allowing colored illumination improves
some PSNR and LPIPS values but lowers SSIM in every setting, reflecting the
extra ambiguity of channel-dependent illumination. We therefore retain the
more constrained achromatic model for stable separation rather than optimizing
one metric through additional decomposition freedom.

\subsection{Real Application: Digital Heritage}
\label{subsec:heritage}

\begin{figure}[thb]
    \centering
    \includegraphics[width=0.32\linewidth]{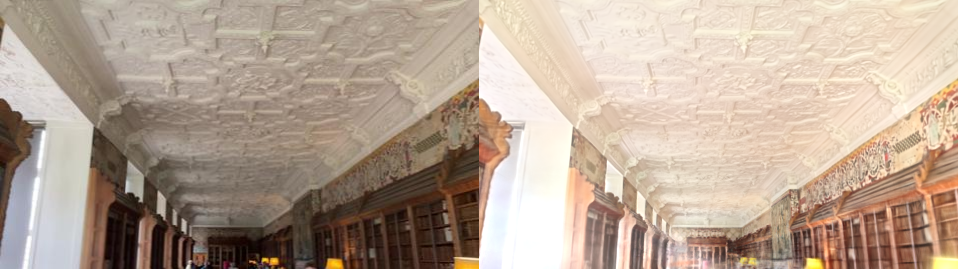}
    \includegraphics[width=0.32\linewidth]{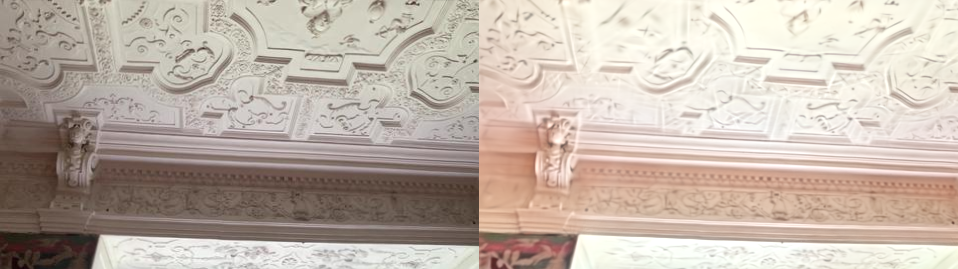}
    \includegraphics[width=0.32\linewidth]{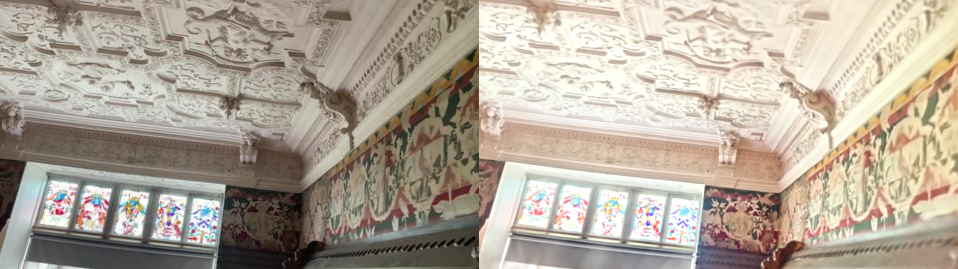}
    \caption{Ceiling digitization under uneven illumination~\citep{ceiling}.
    Each panel pairs a held-out photograph (left) with an EvenSplat rendered novel view
    (right).}
    \label{fig:ceiling}
\end{figure}

Uneven illumination limits the legibility of reconstructed cultural
assets. We apply EvenSplat to ceiling digitization in a large indoor heritage
site~\citep{ceiling}, where windows and local lighting leave reliefs unevenly
exposed. Figure~\ref{fig:ceiling} shows that the resulting novel views reveal
architectural detail across darker regions, illustrating a practical
use beyond the benchmark scenes.

\section{Conclusion}
\label{sec:conclusion}

EvenSplat couples image-space decomposition with shared Gaussian
illumination and global--local view-specific correction. Across physical and
simulated benchmarks, its largest gains occur under spatially uneven and
high-contrast illumination, where global appearance models are insufficient;
ablations confirm that the 2D and 3D branches are
complementary. The result is a practical route to more
illumination-consistent Gaussian scene representations from ordinary sRGB
captures. Limitations include sRGB-domain ambiguity and saturation, the
training-image-specific ILEC parameters, and unmodeled specularities, motion
blur, and sensor noise.

\begingroup
\setlength{\emergencystretch}{3em}
\Urlmuskip=0mu plus 2mu\relax
\bibliography{iclr2027_conference}
\bibliographystyle{plainnat}
\endgroup

\appendix
\section{Appendix}

\input{appendix_losses}

\subsection{Experiment and Implementation Settings}
\label{app:implementation}

This section specifies the training and evaluation settings used throughout
our experiments. Unless stated otherwise, all EvenSplat configurations use the
same loss weights and optimization schedule across different conditions.

\paragraph{Hyperparameters.} Tables~\ref{tab:retinex2d-hyperparams} and~\ref{tab:retinex3d-hyperparams}
report hyperparameters for the image-space decomposition and coupled Gaussian-level settings used by EvenSplat, respectively. 

\begin{table*}[tbh]
\centering
\begin{minipage}[t]{0.49\textwidth}
\centering
\caption{Image-space network settings used across datasets. }
\label{tab:retinex2d-hyperparams}
\begingroup
\scriptsize
\setlength{\tabcolsep}{3pt}
\renewcommand{\arraystretch}{1.05}
\begin{adjustbox}{max width=\linewidth}
\begin{tabular}{lc}
\toprule
\textbf{Hyperparameter} & \textbf{Value} \\
\midrule
\multicolumn{2}{c}{\textit{Loss Weights}} \\
\midrule
$\lambda_{\mathrm{curve}}$ & 0.5 \\
$\lambda_{\mathrm{smooth}}$ & 1.0 \\
$\lambda_{\mathrm{white}}$ & 0.2 \\
\midrule
\multicolumn{2}{c}{\textit{Parameters}} \\
\midrule
$\eta_{\mathrm{dark}}$ & 1.0 \\
$\eta_{\mathrm{bright}}$ & 1.0 \\
$T_{\mathrm{dark}}$ & 0.25 \\
$T_{\mathrm{bright}}$ & 0.65 \\
$\rho_{\mathrm{dark}}$ & 0.45 \\
$\rho_{\mathrm{bright}}$ & 0.55 \\
$\gamma_{\mathrm{edge}}$ & 0.2 \\
$T_{\mathrm{white}}$ & 95.0 \\
$\kappa_{\mathrm{white}}$ & 9.8 \\
$s_{\mathrm{chroma}}$ & 2.4 \\
Image embedding dimension $d_{\mathrm{emb}}$ & 64 \\
\bottomrule
\end{tabular}
\end{adjustbox}
\endgroup
\end{minipage}
\hfill
\begin{minipage}[t]{0.49\textwidth}
\centering
\caption{Coupled Gaussian-level and image correction settings.}
\label{tab:retinex3d-hyperparams}
\begingroup
\scriptsize
\setlength{\tabcolsep}{3pt}
\renewcommand{\arraystretch}{1.05}
\begin{adjustbox}{max width=\linewidth}
\begin{tabular}{lc}
\toprule
\textbf{Hyperparameter} & \textbf{Value} \\
\midrule
\multicolumn{2}{c}{\textit{Loss Weights}} \\
\midrule
$\lambda_{\text{obs}}$ & 1 \\
$\lambda_{\text{excl}}$ & 0.3 \\
$\lambda_{\text{SH}}$ & 0.8 \\
$\lambda_{\text{mag}}$ & 0.1 \\
$\lambda_{\text{shrink}}$ & 0.1 \\
$\lambda_{\text{promote}}$ & 0.1 \\
$\lambda_{\text{recomb}}$ & 0.8 \\
$\lambda_{\text{align}}$ & 0.2 \\
$\lambda_{\text{cpl}}$ & 1 \\
\midrule
\multicolumn{2}{c}{\textit{Parameters}} \\
\midrule
$m_{\mathrm{floor}}$ & 0.15 \\
$\kappa_{\mathrm{guide}}$ & 8 \\
$t_{\mathrm{guide}}$ & 0.45 \\
$\epsilon$ & $10^{-6}$ \\
\bottomrule
\end{tabular}
\end{adjustbox}
\endgroup
\end{minipage}
\end{table*}

\paragraph{Data splits, resolution, and hardware.}
All methods use identical camera poses, image resolutions, and train--test
splits. Views are ordered with every eighth view held out for
evaluation and the remainder used for training. For simulated data, the
perturbed images are training observations and the corresponding clean images
at the same poses are evaluation references.
Simulated scenes are downsampled by a factor of 8 along each spatial dimension, yielding a resolution of approximately $600\times400$ pixels. Real captures are
approximately $3000\times2000$ pixels and are downsampled by a factor of four,
with small scene-dependent variations. RealityScan~2.1 provides fixed camera
poses and a sparse point cloud for initializing the real-world scenes. Training
uses a batch size of one on two NVIDIA GeForce RTX~3090 GPUs with 24~GB memory.

\paragraph{Optimization schedule.}
We optimize each scene for 10,000 iterations with Adam. Gaussian learning rates
are $1.6\times10^{-4}$ for position, $1.4\times10^{-2}$ for scale,
$1.2\times10^{-2}$ for opacity, $1.0\times10^{-3}$ for rotation,
$2.5\times10^{-3}$ for the DC spherical-harmonic (SH) coefficients, and
$1.25\times10^{-4}$ for higher-order SH coefficients. The image-space network
and its appearance embedding use learning rates of $2\times10^{-3}$ and
$5\times10^{-3}$, respectively. CRN, 3D illumination field, and ILEC map use
$3\times10^{-4}$, $6\times10^{-5}$, and $3\times10^{-3}$, respectively.
Position learning rates decay exponentially; the illumination field and CRN
use cosine annealing; and the SH degree increases progressively to degree~3.

\paragraph{Gaussian Illumination Field and Camera Response parameterization.}
The 3D illumination field takes the Gaussian center $\mathbf{x}_j$, Gaussian-to-camera viewing direction $\mathbf{v}_j$, and estimated normal $\mathbf{n}_j$ as input. Positions are scene-normalized and encoded with a 16-level multi-resolution hash grid (2 features per level, base resolution 16, hash size $2^{19}$, per-level scale 1.4473), while $\ell_2$-normalized viewing directions and normals use four-frequency sinusoidal encodings with frequencies $\{1,2,4,8\}$. The concatenated features are processed by a fully fused MLP with hidden width 128 and LeakyReLU activations, followed by a two-layer prediction head whose final layer has zero bias and weights initialized from $\mathcal{N}(0,10^{-4})$. The multiplicative illumination gain is parameterized as
\begin{equation}
\mathbf{a}_j=0.05+3.95\,\sigma(\tilde{\mathbf{a}}_j+\mathbf{1}),
\end{equation}
giving $\mathbf{a}_j\in(0.05,4.0)^3$. 
The current renderer uses only
$\mathbf{c}^{\mathrm{lit}}_j=\mathbf{a}_j\odot\mathbf{c}^{\mathrm{base}}_j$.

Gaussian normals are taken as the rotation axis corresponding to the smallest Gaussian scale and are $\ell_2$-normalized before encoding; no view-dependent normal flipping is applied. The CRN maps each training-image embedding through a width-32 fully fused MLP with LeakyReLU activations and a three-dimensional output head. Its RGB gain is
\begin{equation}
\mathbf{c}^{i}=\operatorname{softplus}(\tilde{\mathbf{c}}^{i}),
\end{equation}
ensuring positivity. The output head uses zero weights, unit gain biases, and zero additive biases, yielding an initial gain of $\operatorname{softplus}(1)\approx1.313$. Gaussian illumination is alpha-composited with the standard transmittance weights without opacity renormalization, and the illumination used for decomposition is clamped to $[0.05,4.0]$ to stabilize low-opacity regions.

\paragraph{Gaussian refinement and staged training.}
Densification and pruning follow the standard 3DGS schedule. Refinement starts
at iteration~500, runs every 100 iterations, and stops at iteration~7,000.
Gaussian growth uses a projected-gradient threshold of $2\times10^{-4}$ and
2D/3D scale thresholds of 0.05/0.01, while pruning uses an opacity threshold
of 0.005 and scale thresholds of 0.15/0.1, with opacity reset every 3,000
iterations. Following Sec.~\ref{subsec:coupling}, the 3D illumination field,
image-space embedding, and CRN are frozen after iteration~3,000. ILEC is enabled at iteration~5,000. All other trainable parameter groups remain active and are updated throughout training.
Each training image has a zero-initialized $\frac{H}{8} \times \frac{W}{8}$ single-channel local log-gain map, bilinearly upsampled, bounded to $[-1.5,1.5]$, and exponentiated, where $H$ and $W$ denote the height and width of the original image.

\paragraph{Novel-view inference and comparison protocol.}
Training-image embeddings and ILEC maps are image specific and are not assigned
to held-out views. At inference, the CRN is replaced by the identity response
and the ILEC gain is fixed to one; novel-view rendering therefore relies only
on the shared Gaussian representation and the learned 3D illumination field.
All methods are trained and evaluated on the same input data and paired
held-out views. PSNR, SSIM, and LPIPS are computed directly between rendered
images and their corresponding references. Unless otherwise stated, no
test-time pose refinement, exposure fitting, affine RGB correction, or color
matching is applied; corrected metrics are retained only as diagnostics and
are excluded from the primary comparisons.

\paragraph{Baseline configuration and test-view inference.}

We follow the publicly released implementations and their native inference
protocols. For 3DGS, we use the default gsplat configuration. For 3DGS+CHROMA, we use the official CHROMA implementation to harmonize the training images in batches of up to 500 images, and then train 3DGS on the corrected images using the same default gsplat configuration. For GS-W, we enable its per-image appearance optimization during training and render
held-out views from the learned intrinsic/base appearance representation,
without optimizing an additional appearance code for the test image. For
Bilateral Grid, per-image bilateral grids are used only during training to
absorb view-specific photometric discrepancies; the correction branch is
bypassed for held-out views, which are rendered directly from the learned 3D
representation. Luminance-GS is evaluated with its released view-adaptive
configuration, while PPISP uses its released post-processing pipeline together
with the MCMC Gaussian strategy. We do not introduce zero, mean, nearest-view,
or other appearance-code heuristics beyond those defined by the corresponding
released methods.

\paragraph{Tone-Matched 3DGS baseline.}

Tone-matched 3DGS is a control baseline that applies a fixed global exposure correction
to vanilla 3DGS renderings. For each rendered RGB image
$\mathbf{I}=(R,G,B)$, we first compute its per-pixel luminance as
$Y=0.2126R+0.7152G+0.0722B$, and denote by $\bar{Y}$ the mean luminance
averaged over all pixels in the rendered image. We then determine a global
scalar gain
$\alpha=0.5/(\bar{Y}+\epsilon)$
and rescale the rendered image as
$\mathbf{I}'=\operatorname{clip}(\alpha\mathbf{I},0,1)$,
where $0.5$ is a fixed prescribed mid-tone target and $\epsilon$ is a small
constant for numerical stability. This operation does not use ground-truth
images and introduces neither additional learnable parameters nor test-time
optimization. It serves as a control for quantifying how much improvement can
be obtained from simple global tone/exposure normalization alone, thereby
separating gains due to overall brightness adjustment from those provided by
our image-space decomposition, 3D illumination modeling, and local exposure compensation.

\paragraph{Directional-lighting evaluation.}
The paired Robot experiment measures appearance consistency under two
controlled lighting directions. The room illumination is disabled and the
camera follows a fixed trajectory around the robot. At each camera position,
we capture one image with a left-side light and one with a right-side light,
without changing the camera pose or settings. This produces 74 matched view
pairs with shared camera parameters and poses.

The two lighting datasets are fitted independently, with no learned scene
parameters shared between them. All 74 views are used for fitting and for the
paired-view comparison. This experiment therefore measures consistency at
observed, pose-matched views rather than held-out novel-view generalization.
For each pose, we compare the two appearance renderings over the full image,
without foreground masking or spatial cropping. 

PSNR is computed independently for each matched pair and then averaged:
\begin{equation}
\overline{\mathrm{PSNR}}
=\frac{1}{74}\sum_{i=1}^{74}
\mathrm{PSNR}\!\left(
\hat{\mathbf{R}}_{\mathrm{L}}^{i},
\hat{\mathbf{R}}_{\mathrm{R}}^{i}
\right),
\end{equation}
where $\hat{\mathbf{R}}_{\mathrm{L}}^{i}$ and
$\hat{\mathbf{R}}_{\mathrm{R}}^{i}$ are appearances at the same pose from the
models fitted to the left- and right-lighting datasets. With RGB values in
$[0,1]$, PSNR uses unit peak intensity. The reported score is the arithmetic
mean of the per-view values rather than a score computed from a pooled MSE.

\subsection{Dataset Construction and Capture Details}
\label{app:capture}
This section gives the construction details omitted from the main paper.
The real benchmark contains nine scenes acquired under controlled exposure
and illumination variation, while the simulated benchmark applies three
perturbation families to six Mip-NeRF~360~\citep{barron2022mipnerf360} scenes.

\subsubsection{Real-world Capture Datasets}
\label{app:capture_real}

\paragraph{Capture equipment and general protocol.}
The real benchmark contains three CEV, four SIV, and two HCI scenes, spanning
indoor and outdoor subjects. Table~\ref{tab:capture-details} records 78--472
reference views per scene and all captured exposure or illumination variants.
We use a Canon EOS 700D with an 18--135~mm f/3.5--5.6 lens and a Sony
$\alpha$7 III with a 24--105~mm f/4.0 lens. All scenes are captured handheld
to accommodate uneven terrain and viewpoints inaccessible to a tripod.
Automatic exposure bracketing and high-rate burst capture reduce motion
between consecutive variants, although small residual misalignment can remain.

\paragraph{Real CEV capture.}
Exposure bracketing changes brightness between views. On the Canon EOS 700D,
the Magic Lantern add-on extends the available bracket from the camera's
default $\pm2$~EV to $\pm5$~EV, creating a wider range of cross-view exposure
variation.

\paragraph{Real SIV capture.}
Localized illumination is produced with an external flash partly obstructed
by aluminum foil. For most SIV scenes, the first burst image is deliberately
underexposed while the flash fires at maximum power; subsequent reference
images are captured without flash while it recharges. For
\textit{PlasticCart}, flash--shutter mis-synchronization creates a sharper
illumination boundary across the image.

\paragraph{Real HCI capture.}
Close-range flash, partial obstruction, and low camera exposure create a
brightly illuminated region surrounded by deep shadows. The first images in
the burst are heavily underexposed, yielding both saturated highlights and
severely dark regions. The capture log identifies Canon for \textit{LogHC}
and Sony for \textit{RobotHC}.

\paragraph{Reference capture and reconstruction inputs.}
Reference photographs use nominal 0.0~EV. Outdoor references are captured
under overcast skies and indoor references under uniform room lighting to
reduce directional shadows and spatial imbalance. Canon references use
evaluative metering; per-scene ISO, aperture, and shutter settings are listed
in Table~\ref{tab:capture-details}. These photographs remain sRGB appearance
references containing camera response and residual scene illumination, rather
than calibrated reflectance. RealityScan 2.1 estimates camera alignment and
sparse points for Gaussian initialization.

\begin{table*}[!t]
\centering
\caption{Capture log for the nine real-world evaluation scenes. Reference
views denote the number of reference photographs; All Images gives the total
including captured exposure and illumination variants. CEV, SIV, and HCI denote the three
illumination settings defined in Sec.~\ref{subsec:datasets}. AUTO denotes
an automatically selected camera setting; shutter speeds are in seconds.
Canon and Sony denote the EOS 700D and $\alpha$7 III, respectively.}
\label{tab:capture-details}
\scriptsize
\begin{adjustbox}{max width=\textwidth}
\begin{tabular}{cccccccc}
\toprule
\textbf{Dataset Scene} & \textbf{Reference Views} & \textbf{All Images} & \textbf{ISO} & \textbf{f/stop} & \textbf{Shutter Speed} & \textbf{Category} & \textbf{Camera Model} \\
\hline
LogHC & 122 & 244 & AUTO & f4 & 1/200 & HCI & Canon \\
RobotHC & 183 & 366 & AUTO & f10 & 1/200 & HCI & Sony \\
FourLogs & 211 & 633 & AUTO & f3.5 & AUTO & CEV & Canon \\
Log & 260 & 780 & AUTO & f3.5 & AUTO & CEV & Canon \\
Robot & 472 & 1416 & AUTO & f5.6 & AUTO & CEV & Canon \\
ParkBench & 124 & 248 & AUTO & f4 & 1/200 & SIV & Sony \\
CircularBench & 78 & 156 & AUTO & f5 & 1/200 & SIV & Sony \\
PlasticCart & 123 & 492 & 6400 & f5.6 & AUTO & SIV & Sony \\
Rocks & 120 & 240 & AUTO & f7.1 & 1/200 & SIV & Sony \\
\bottomrule
\end{tabular}
\end{adjustbox}

\end{table*}
\FloatBarrier

\subsubsection{Simulated Datasets}
\label{app:capture_simulated}

\paragraph{Simulated scenes and references.}
The quantitative benchmark uses \textit{Counter}, \textit{Bonsai},
\textit{Bicycle}, \textit{Garden}, \textit{Kitchen}, and \textit{Room} from
Mip-NeRF~360~\citep{barron2022mipnerf360}. Perturbations are applied only to
the training images; the original images are retained as evaluation
references. Each setting therefore measures recovery toward the same
unmodified scene appearance.

\paragraph{Simulated CEV.}
We use the exposure-variation benchmark introduced by
\citet{cui_luminance_gs} (their \texttt{variance} dataset), which assigns
different exposure levels to training views and applies a mild gamma
adjustment. This produces view-to-view photometric inconsistency while
preserving the scene content and camera trajectory.

\paragraph{Simulated SIV.}
For each training image, low and high multiplicative exposure factors are
sampled from $[0.4,0.6]$ and $[1.4,1.6]$. The image is divided horizontally
or vertically at a position sampled between 30\% and 70\% of the corresponding
dimension. The two sides receive opposite factors; a Gaussian-smoothed binary
mask blends the regions into a gradual transition. The result is clipped to
the valid image range.

\paragraph{Simulated HCI.}
We increase training-image contrast by a factor of three and clip the result
to the valid image range, producing stronger bright--dark imbalance than the
SIV perturbation. 

\subsection{Supplementary Results} \label{app:simulated_results}

\paragraph{ILEC regularization ablation.} 

ILEC models image-specific spatial exposure residuals that remain after
Gaussian-level illumination modeling. The module-level ablation in
Table~\ref{tab:retinex3d-ablation} shows its largest PSNR benefit under high
contrast, where a single global correction cannot absorb
the residual bright--dark imbalance. We therefore examine its regularization terms in this setting in
Table~\ref{tab:ILEC-ablation}. It shows that
all three constraints matter: the magnitude penalty limits arbitrary local
edits, exposure promotion recovers persistently dark regions, and
guidance-weighted shrinkage provides the strongest contribution (a 0.62~dB
PSNR drop when removed). The result supports a constrained, illumination-aware
local model rather than an unconstrained per-image correction map.

\begin{table}[H]
\setlength{\tabcolsep}{8pt}
\renewcommand{\arraystretch}{1.2}
\caption{ILEC ablations on real-world High-Contrast Illumination (HCI) dataset. }
\label{tab:ILEC-ablation}
\centering
\begin{adjustbox}{max width=\columnwidth}
\begin{tabular}{lc}
\hline\hline
    \textbf{Variant} & PSNR/SSIM/LPIPS \\
\hline
Full            & 16.950/0.682/0.241 \\
No Magnitude Penalty  & 16.723/0.653/0.243 \\
No Guided Shrinkage& 16.330/0.647/0.276 \\
No Exposure Promotion  & 16.600/0.668/0.260 \\
\hline\hline
\end{tabular}
\end{adjustbox}
\end{table}

\paragraph{Additional results on real-world dataset.}

\begin{figure*}[h]
    \centering
    \includegraphics[width=0.95\textwidth, keepaspectratio]{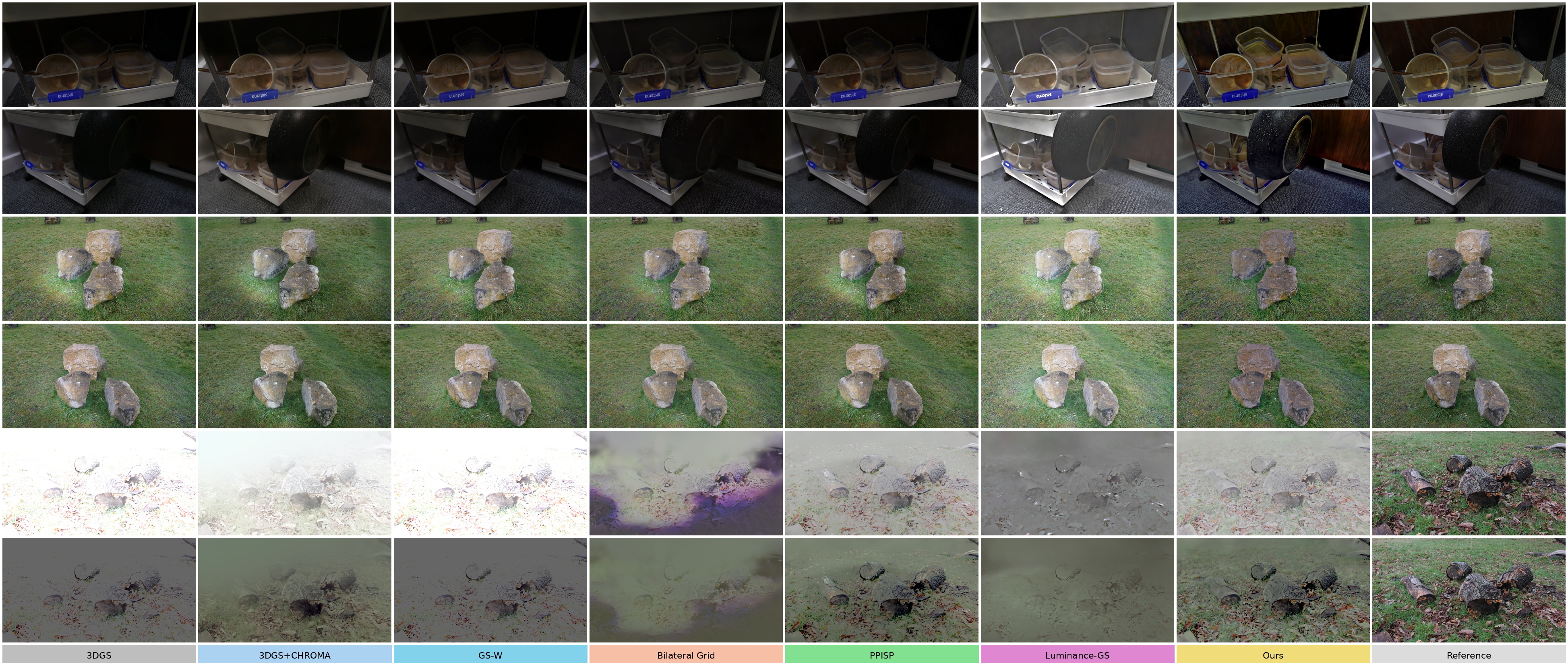}
    \caption{
    Additional real-world qualitative comparisons. Columns show 3DGS, 3DGS+CHROMA, GS-W,
    Bilateral Grid, PPISP, Luminance-GS, EvenSplat, and the reference. The first
    two rows show \textit{PlasticCart} (SIV), the next two show \textit{Rocks}
    (SIV), and the final two show the extreme \textit{FourLogs} case (CEV).
    For \textit{FourLogs}, the last row applies a per-image, per-channel affine
    correction, $y_c=a_cx_c+b_c$, to the preceding outputs. This diagnostic
    removes global color mismatch and exposes the remaining spatial
    reconstruction artifacts.
    }
    \label{fig:real-world}
\end{figure*}

Table~\ref{tab:all-results} expands the main comparison to all nine scenes.
The breakdown confirms that the HCI gain occurs on both high-contrast scenes
and that the aggregate SIV improvement is not attributable to one example.
It also exposes the expected CEV variability, where methods specialized for
global photometric correction can remain competitive. Figure~\ref{fig:real-world}
adds difficult cases and separates global color mismatch from spatial
illumination error through an affine-corrected visualization for the extremely exposed \emph{FourLogs} scene.

\begin{table*}[h]
\centering
\caption{
Per-scene quantitative comparisons on the real-world dataset under CEV, SIV,
and HCI. Each entry reports
PSNR$\uparrow$/SSIM$\uparrow$/LPIPS$\downarrow$.
\textbf{Tone-Matched 3DGS} applies a single global tone/exposure normalization
to vanilla 3DGS to match the prescribed mid-tone target of 0.5. This control
isolates the contribution of global tone matching from the proposed
decomposition. Best results are highlighted in light red and second-best
results in light yellow.
}
\label{tab:all-results}
\begingroup
\fontsize{7}{8}\selectfont
\setlength{\tabcolsep}{3pt}
\setlength{\fboxsep}{0.7pt}
\renewcommand{\arraystretch}{1.16}
\newcommand{\allbest}[1]{\colorbox{bestcell}{\textbf{#1}}}
\newcommand{\allsecond}[1]{\colorbox{secondcell}{\underline{#1}}}
\begin{adjustbox}{width=\textwidth}
\begin{tabular}{@{}llcccccccc@{}}
\toprule
\textbf{Setting} & \textbf{Scene}  & \textbf{3DGS} & \textbf{3DGS+CHROMA} & \textbf{Bilateral Grid}
& \textbf{GS-W} & \textbf{Luminance-GS} & \textbf{PPISP}
& \textbf{EvenSplat} & \textbf{Tone-Matched 3DGS} \\
\midrule

\multirow{3}{*}{\textbf{CEV}}
& \textit{FourLogs}
& 10.954/0.131/0.940
& 7.017/0.376/0.680 
& 12.339/0.186/0.928
& 8.965/0.094/\allsecond{0.641}
& \allbest{15.684}/0.223/0.867
& 11.006/\allbest{0.436}/\allbest{0.560}
& \allsecond{15.148}/\allsecond{0.427}/0.662
& 13.035/0.177/0.952 \\

& \textit{Log}
& 17.072/0.651/0.367
& 10.503/0.547/0.294
& 20.130/0.783/0.216
& 13.432/0.676/0.220
& 17.745/0.734/0.283
& \allsecond{20.547}/\allbest{0.871}/\allbest{0.115}
& \allbest{20.648}/\allsecond{0.812}/\allsecond{0.157}
& 17.675/0.662/0.321 \\

& \textit{Robot}
& 14.564/0.601/0.519
& 13.060/0.666/0.271
& 16.542/0.677/0.429
& 14.814/0.798/0.224
& 16.325/0.656/0.304
& \allsecond{21.829}/\allsecond{0.861}/\allsecond{0.133}
& \allbest{22.781}/\allbest{0.884}/\allbest{0.097}
& 14.117/0.678/0.393 \\

\midrule

\multirow{4}{*}{\textbf{SIV}}
& \textit{PlasticCart}
& 14.754/0.595/0.498
& 10.597/0.380/0.350\
& 15.096/0.651/0.415
& 14.549/0.586/0.402
& 14.456/0.645/0.400
& \allsecond{15.826}/\allsecond{0.683}/\allsecond{0.258}
& \allbest{18.427}/\allbest{0.750}/\allbest{0.253}
& 10.272/0.581/0.513 \\

& \textit{Rocks}
& 20.591/0.827/\allsecond{0.121}
& \allbest{21.248}/0.802/0.164
& 20.825/0.826/0.122
& 20.243/0.805/0.144
& 14.599/0.750/0.171
& 20.956/\allbest{0.842}/\allsecond{0.121}
& \allsecond{21.162}/\allsecond{0.838}/\allbest{0.118}
& 15.152/0.765/0.156 \\

& \textit{ParkBench}
& \allsecond{20.310}/\allsecond{0.810}/\allsecond{0.135}
& 20.828/0.785/0.165
& 19.907/0.803/0.138
& 20.043/0.761/0.168
& 18.248/0.803/0.136
& 19.810/\allbest{0.816}/0.142
& \allbest{21.037}/\allbest{0.816}/\allbest{0.117}
& 18.002/0.787/0.146 \\

& \textit{CircularBench}
& 15.953/0.702/0.244
& 15.056/0.661/0.275
& \allsecond{16.147}/0.690/0.233
& 16.134/0.654/0.257
& 14.928/0.725/0.214
& 15.854/\allbest{0.741}/\allsecond{0.191}
& \allbest{17.420}/\allsecond{0.739}/\allbest{0.186}
& 14.804/0.671/0.255 \\

\midrule

\multirow{2}{*}{\textbf{HCI}}
& \textit{LogHC}
& 10.533/0.393/0.328
& 10.597/0.380/0.350
& 11.312/0.438/0.302
& 10.686/0.396/0.325
& \allsecond{12.614}/\allsecond{0.565}/\allsecond{0.287}
& 10.568/0.400/0.322
& \allbest{16.718}/\allbest{0.669}/\allbest{0.220}
& 10.142/0.542/0.315
 \\

& \textit{RobotHC}
& 12.870/0.460/0.452
& \allsecond{16.963}/\allsecond{0.664}/\allsecond{0.353}
& 13.068/0.370/0.395
& 12.916/0.454/0.432
& 14.468/0.540/0.394
& 13.307/0.491/0.367
& \allbest{17.181}/\allbest{0.694}/\allbest{0.262}
& 11.737/0.603/0.419\\

\bottomrule
\end{tabular}
\end{adjustbox}
\endgroup
\end{table*}

\paragraph{Additional results on simulated dataset.}

\begin{figure*}[htb]
    \centering
    \includegraphics[width=0.95\linewidth]{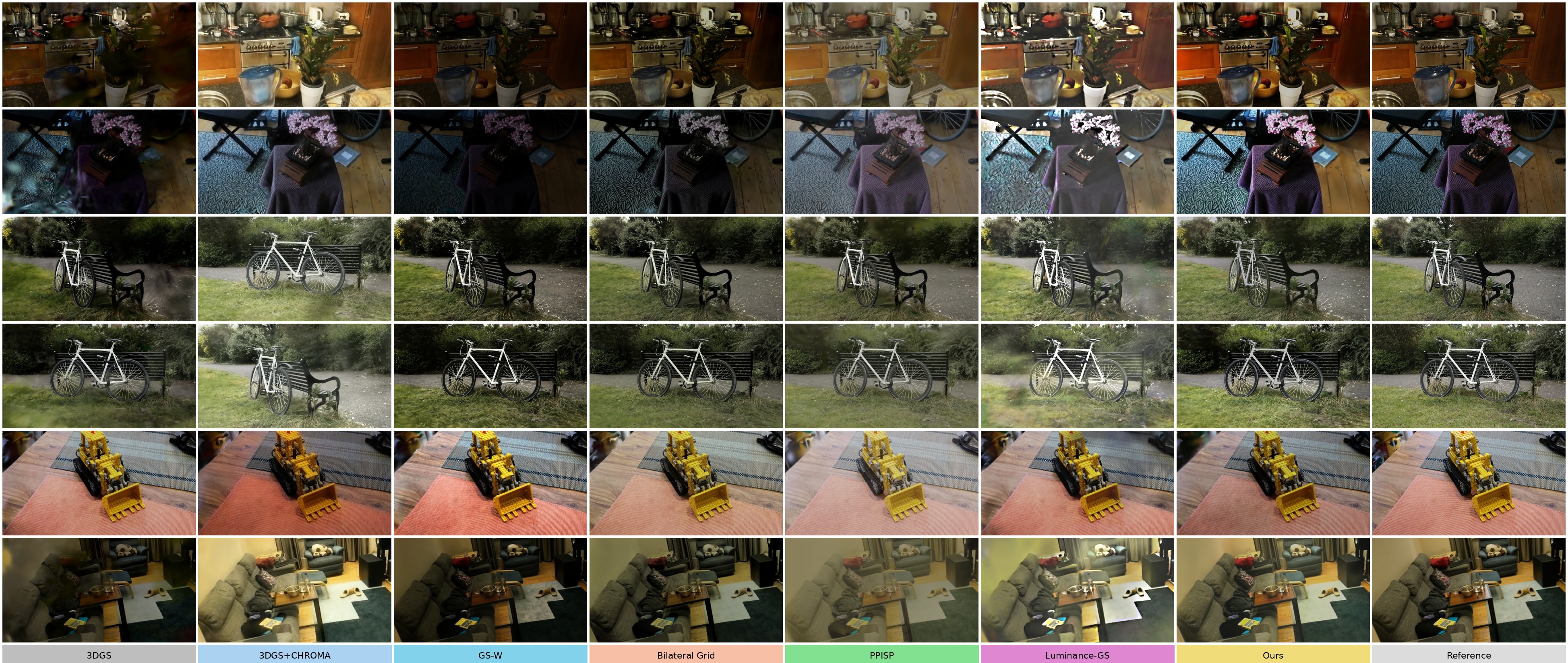}
    \caption{Qualitative comparison on simulated scenes under Cross-View
    Exposure Variation (CEV). Columns show 3DGS, 3DGS+CHROMA, GS-W, Bilateral Grid, PPISP,
    Luminance-GS, EvenSplat, and the reference; each row is a held-out view from
    a different scene. CEV changes the global exposure between views, testing
    whether each method can recover a consistent scene appearance without
    retaining view-dependent brightness and color shifts.}
    \label{fig:comp-variance-supp}
\end{figure*}

\begin{figure*}[htb]
    \centering
    \includegraphics[width=0.95\linewidth]{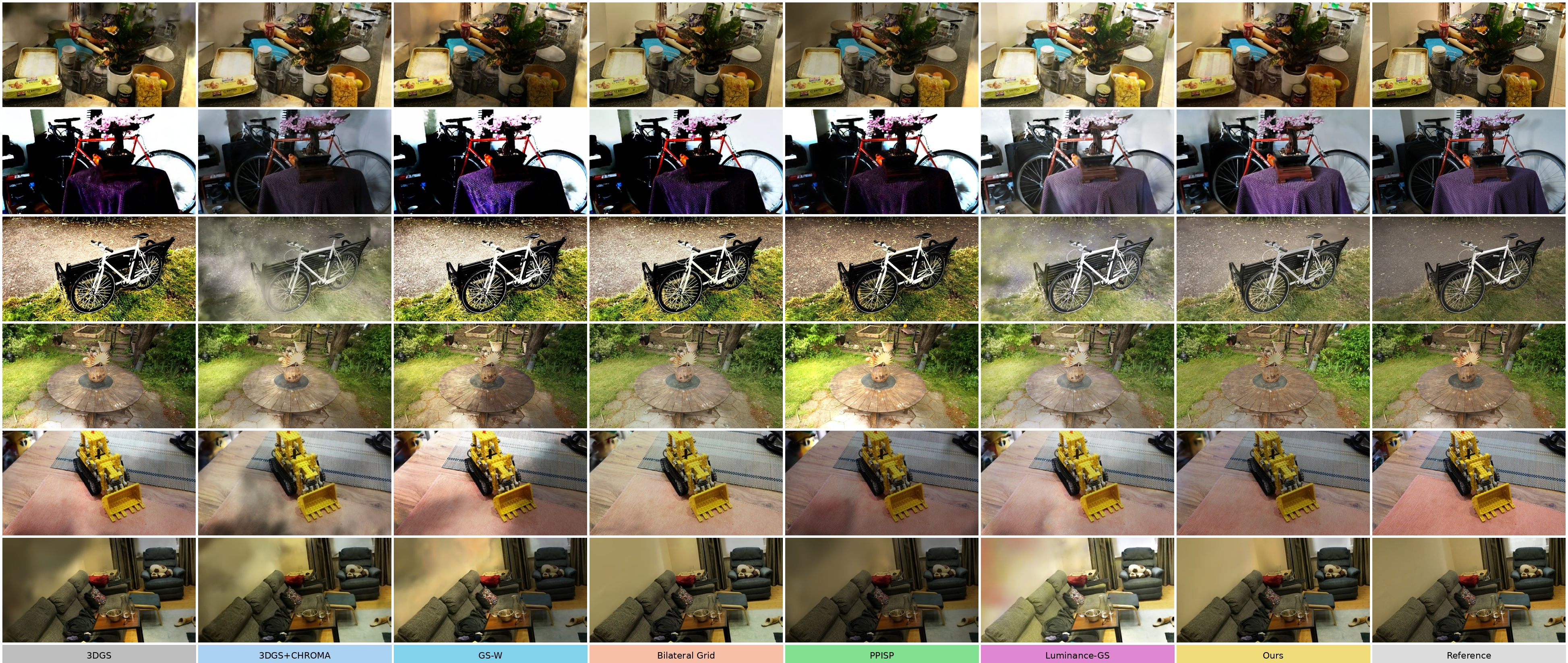}
    \caption{Qualitative comparison on simulated scenes under Spatial
    Illumination Variation (SIV). Columns show 3DGS, 3DGS+CHROMA, GS-W, Bilateral Grid,
    PPISP, Luminance-GS, EvenSplat, and the reference; each row is a held-out
    view from a different scene. The illumination changes within each image,
    producing adjacent bright and dark regions that cannot be removed by a
    single global exposure correction.}
    \label{fig:comp-multiexposure-supp}
\end{figure*}

\begin{figure*}[htb]
    \centering
    \includegraphics[width=0.95\linewidth]{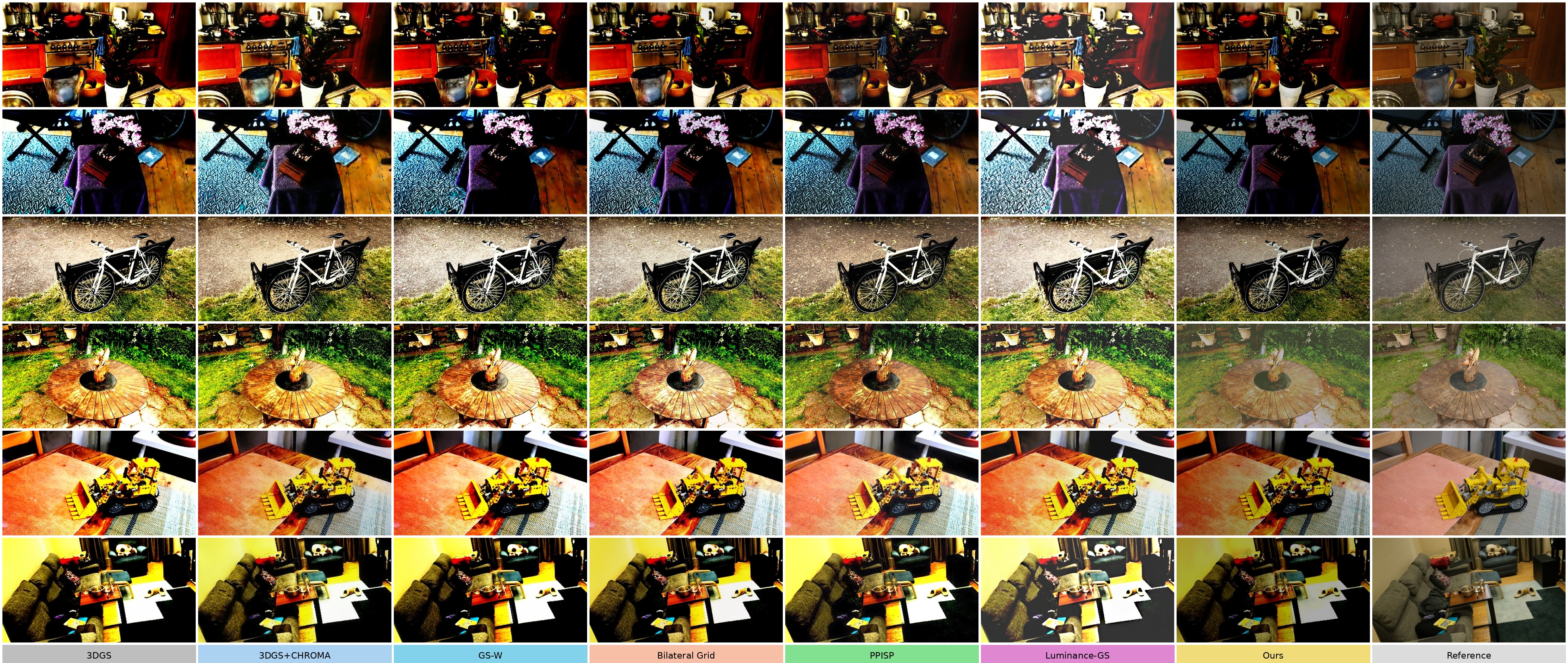}
    \caption{Qualitative comparison on simulated scenes under High-Contrast
    Illumination (HCI). Columns show 3DGS, 3DGS+CHROMA, GS-W, Bilateral Grid, PPISP,
    Luminance-GS, EvenSplat, and the reference; each row is a held-out view from
    a different scene. Strong highlights and deep shadows create clipping and
    severe local imbalance, emphasizing whether a method can restore shadow
    detail without flattening or overexposing bright regions.}
    \label{fig:comp-contrast-supp}
\end{figure*}

Table~\ref{tab:simulated-per-scene} shows that the aggregate CEV and SIV gains
extend across scenes with different geometry and texture. HCI remains more
metric-dependent, consistent with the main results: correcting clipped,
high-contrast inputs improves intensity agreement but can alter local structure
and perceptual similarity. Figures~\ref{fig:comp-variance-supp}--\ref{fig:comp-contrast-supp} visualize the same pattern, with the clearest
advantage near exposure transitions and in shadowed regions.

\begin{table}[h]
\caption{Per-scene quantitative comparisons on simulated data under
(a) Cross-View Exposure Variation (CEV), (b) Spatial Illumination Variation
(SIV), and (c) High-Contrast Illumination (HCI). Each panel reports the six
scenes. Higher PSNR and
SSIM and lower LPIPS are better. Following Table~\ref{tab:all-results},
best values are bold with light-red shading and second-best values are
underlined with light-yellow shading, ranked separately for each scene and
metric. Ties at the reported precision share the same rank.}
\label{tab:simulated-per-scene}
\centering
\begin{minipage}{\textwidth}
\centering
\textbf{(a) CEV}\par\smallskip
\begingroup
\fontsize{6}{7}\selectfont
\setlength{\tabcolsep}{1.2pt}
\renewcommand{\arraystretch}{1.12}
\begin{adjustbox}{max width=\linewidth}
\begin{tabular}{l*{18}{r}}
\toprule
\textbf{Method} & \multicolumn{3}{c}{\textbf{Counter}} & \multicolumn{3}{c}{\textbf{Bonsai}} & \multicolumn{3}{c}{\textbf{Bicycle}} & \multicolumn{3}{c}{\textbf{Garden}} & \multicolumn{3}{c}{\textbf{Kitchen}} & \multicolumn{3}{c}{\textbf{Room}} \\
\cmidrule(lr){2-4}\cmidrule(lr){5-7}\cmidrule(lr){8-10}\cmidrule(lr){11-13}\cmidrule(lr){14-16}\cmidrule(lr){17-19}
 & PSNR$\uparrow$ & SSIM$\uparrow$ & LPIPS$\downarrow$ & PSNR$\uparrow$ & SSIM$\uparrow$ & LPIPS$\downarrow$ & PSNR$\uparrow$ & SSIM$\uparrow$ & LPIPS$\downarrow$ & PSNR$\uparrow$ & SSIM$\uparrow$ & LPIPS$\downarrow$ & PSNR$\uparrow$ & SSIM$\uparrow$ & LPIPS$\downarrow$ & PSNR$\uparrow$ & SSIM$\uparrow$ & LPIPS$\downarrow$ \\
\midrule
3DGS & 14.1813 & 0.5011 & 0.3741 & 12.7882 & 0.3272 & 0.5585 & 19.0437 & 0.5934 & 0.3449 & 20.1326 & 0.7720 & 0.1702 & 22.9005 & 0.8686 & \second{0.0894} & 14.6840 & 0.5686 & 0.3353 \\
3DGS+CHROMA & 15.2677 & \second{0.7024} & \second{0.1928} & \best{22.0536} & \second{0.7860} & \second{0.2160} & 14.2583 & 0.6413 & 0.2882 & \second{22.2103} & \best{0.8312} & \second{0.1588} & 15.6871 & 0.7849 & 0.1373 & 15.9184 & 0.7532 & 0.1974 \\
Bilateral Grid & \second{17.2331} & 0.5907 & 0.2759 & 14.2005 & 0.4373 & 0.4298 & 19.7586 & 0.6825 & 0.3256 & 20.0180 & 0.7032 & 0.1812 & 21.7525 & 0.7769 & 0.1228 & \second{19.6984} & \best{0.8521} & \second{0.1842} \\
GS-W & 14.7554 & 0.5824 & 0.2077 & 13.1171 & 0.3968 & 0.3191 & \second{21.2405} & 0.6334 & 0.2879 & 18.9876 & 0.7605 & 0.1604 & 21.1178 & 0.8408 & 0.1110 & 15.8833 & 0.6255 & 0.2018 \\
Luminance-GS & 15.5841 & 0.6295 & 0.2896 & 15.2206 & 0.5636 & 0.3953 & 16.0880 & 0.6237 & 0.3342 & 19.5610 & 0.7976 & 0.1832 & \second{23.6768} & \second{0.8774} & 0.1024 & 15.0968 & 0.6374 & 0.3098 \\
PPISP & 16.4238 & 0.6399 & 0.2238 & 17.8789 & 0.6774 & 0.2163 & \best{22.7584} & \best{0.7108} & \second{0.2711} & 21.0851 & 0.7907 & 0.1709 & 18.6551 & 0.8269 & 0.1031 & 18.5316 & 0.7076 & 0.2110 \\
EvenSplat & \best{18.7079} & \best{0.7644} & \best{0.1638} & \second{18.2206} & \best{0.7870} & \best{0.1493} & 20.3733 & \second{0.6901} & \best{0.2583} & \best{23.6465} & \second{0.8310} & \best{0.1362} & \best{24.6580} & \best{0.8806} & \best{0.0879} & \best{20.2534} & \second{0.8242} & \best{0.1548} \\
\bottomrule
\end{tabular}
\end{adjustbox}
\endgroup
\end{minipage}
\par\medskip
\begin{minipage}{\textwidth}
\centering
\textbf{(b) SIV}\par\smallskip
\begingroup
\fontsize{6}{7}\selectfont
\setlength{\tabcolsep}{1.2pt}
\renewcommand{\arraystretch}{1.12}
\begin{adjustbox}{max width=\linewidth}
\begin{tabular}{l*{18}{r}}
\toprule
\textbf{Method} & \multicolumn{3}{c}{\textbf{Counter}} & \multicolumn{3}{c}{\textbf{Bonsai}} & \multicolumn{3}{c}{\textbf{Bicycle}} & \multicolumn{3}{c}{\textbf{Garden}} & \multicolumn{3}{c}{\textbf{Kitchen}} & \multicolumn{3}{c}{\textbf{Room}} \\
\cmidrule(lr){2-4}\cmidrule(lr){5-7}\cmidrule(lr){8-10}\cmidrule(lr){11-13}\cmidrule(lr){14-16}\cmidrule(lr){17-19}
 & PSNR$\uparrow$ & SSIM$\uparrow$ & LPIPS$\downarrow$ & PSNR$\uparrow$ & SSIM$\uparrow$ & LPIPS$\downarrow$ & PSNR$\uparrow$ & SSIM$\uparrow$ & LPIPS$\downarrow$ & PSNR$\uparrow$ & SSIM$\uparrow$ & LPIPS$\downarrow$ & PSNR$\uparrow$ & SSIM$\uparrow$ & LPIPS$\downarrow$ & PSNR$\uparrow$ & SSIM$\uparrow$ & LPIPS$\downarrow$ \\
\midrule
3DGS & 19.0273 & 0.7834 & 0.1614 & 19.5389 & 0.8409 & 0.1322 & 18.4637 & 0.6237 & 0.3472 & 17.2341 & 0.7695 & 0.1732 & 16.3484 & 0.8043 & 0.1822 & 18.8131 & 0.8171 & 0.1605 \\
3DGS+CHROMA & 19.6773 & 0.7838 & 0.1680 & 19.9524 & 0.7847 & 0.1811 & 16.5234 & 0.6110 & 0.3727 & 18.7528 & 0.7441 & 0.2533 & 17.3001 & 0.7658 & 0.2436 & 18.7915 & 0.7785 & 0.2233 \\
Bilateral Grid & 19.1600 & 0.7443 & 0.1515 & 20.2708 & 0.8680 & 0.1200 & 18.3092 & \second{0.7358} & \second{0.1392} & \second{19.6960} & 0.7493 & 0.1637 & 18.6821 & 0.7593 & 0.1550 & 18.4933 & 0.7809 & 0.1325 \\
GS-W & 19.9049 & 0.7846 & 0.1450 & 20.3154 & 0.8532 & 0.1072 & \second{19.1688} & 0.6525 & 0.2876 & 17.4411 & 0.7647 & 0.1746 & 15.7612 & 0.8046 & 0.1484 & 17.6349 & 0.7957 & 0.1775 \\
Luminance-GS & 13.6157 & 0.6462 & 0.2126 & 12.7315 & 0.6407 & 0.1978 & 13.1440 & 0.5931 & 0.3384 & 16.4130 & 0.7530 & 0.2180 & 18.2343 & 0.8309 & 0.1552 & 12.7719 & 0.6265 & 0.2540 \\
PPISP & \second{20.8596} & \second{0.8238} & \second{0.1270} & \second{22.2038} & \second{0.8879} & \second{0.0911} & 18.3908 & 0.6482 & 0.3030 & \best{20.3391} & \second{0.8074} & \second{0.1508} & \second{19.5526} & \second{0.8534} & \second{0.1122} & \second{20.7150} & \second{0.8604} & \second{0.1236} \\
EvenSplat & \best{21.7063} & \best{0.8408} & \best{0.0976} & \best{23.8880} & \best{0.9064} & \best{0.0669} & \best{19.7933} & \best{0.8211} & \best{0.1146} & 19.4974 & \best{0.8090} & \best{0.1370} & \best{22.5320} & \best{0.8956} & \best{0.0850} & \best{24.5839} & \best{0.8965} & \best{0.0802} \\
\bottomrule
\end{tabular}
\end{adjustbox}
\endgroup
\end{minipage}
\par\medskip
\begin{minipage}{\textwidth}
\centering
\textbf{(c) HCI}\par\smallskip
\begingroup
\fontsize{6}{7}\selectfont
\setlength{\tabcolsep}{1.2pt}
\renewcommand{\arraystretch}{1.12}
\begin{adjustbox}{max width=\linewidth}
\begin{tabular}{l*{18}{r}}
\toprule
\textbf{Method} & \multicolumn{3}{c}{\textbf{Counter}} & \multicolumn{3}{c}{\textbf{Bonsai}} & \multicolumn{3}{c}{\textbf{Bicycle}} & \multicolumn{3}{c}{\textbf{Garden}} & \multicolumn{3}{c}{\textbf{Kitchen}} & \multicolumn{3}{c}{\textbf{Room}} \\
\cmidrule(lr){2-4}\cmidrule(lr){5-7}\cmidrule(lr){8-10}\cmidrule(lr){11-13}\cmidrule(lr){14-16}\cmidrule(lr){17-19}
 & PSNR$\uparrow$ & SSIM$\uparrow$ & LPIPS$\downarrow$ & PSNR$\uparrow$ & SSIM$\uparrow$ & LPIPS$\downarrow$ & PSNR$\uparrow$ & SSIM$\uparrow$ & LPIPS$\downarrow$ & PSNR$\uparrow$ & SSIM$\uparrow$ & LPIPS$\downarrow$ & PSNR$\uparrow$ & SSIM$\uparrow$ & LPIPS$\downarrow$ & PSNR$\uparrow$ & SSIM$\uparrow$ & LPIPS$\downarrow$ \\
\midrule
3DGS & 13.0949 & 0.3921 & 0.3928 & 12.5985 & 0.4295 & 0.3489 & 13.2249 & 0.4210 & 0.3453 & 13.0387 & 0.4530 & 0.3874 & 13.5651 & 0.5538 & 0.3216 & 12.4804 & 0.4633 & 0.3537 \\
3DGS+CHROMA & 13.9290 & \second{0.4788} & \best{0.3291} & 13.2153 & 0.4934 & 0.3286 & 14.4233 & 0.4257 & 0.3798 & 13.5515 & 0.4803 & 0.3837 & \best{16.1563} & 0.6362 & 0.2870 & \second{13.9091} & 0.4995 & 0.3356 \\
Bilateral Grid & 14.0340 & 0.4497 & 0.3523 & 13.8610 & 0.4642 & 0.3070 & 14.2752 & 0.4526 & 0.3141 & 14.0389 & 0.5104 & 0.3583 & 14.4848 & 0.6033 & 0.2833 & 13.6416 & 0.4970 & 0.3195 \\
GS-W & 12.9158 & 0.3906 & 0.3846 & 12.7446 & 0.4391 & 0.3444 & 13.2339 & 0.4137 & 0.3528 & 12.9188 & 0.4449 & 0.3909 & 13.6051 & 0.5441 & 0.3263 & 12.5066 & 0.4711 & 0.3602 \\
Luminance-GS & 12.3740 & 0.4502 & 0.3729 & 11.3973 & \second{0.5395} & 0.3541 & 12.2760 & \second{0.5627} & 0.3376 & 14.7142 & \best{0.6318} & 0.3548 & 15.3304 & \second{0.6451} & 0.3187 & 11.6883 & \second{0.5757} & 0.3468 \\
PPISP & \best{17.2665} & \best{0.5271} & 0.3518 & \second{14.6068} & \best{0.5727} & \second{0.2930} & \second{15.1860} & \best{0.5831} & \best{0.2864} & \best{15.3973} & \second{0.5956} & \second{0.3254} & \second{15.6006} & \best{0.6809} & \best{0.2404} & 12.9865 & \best{0.5767} & \second{0.3170} \\
EvenSplat & \second{16.6347} & 0.4659 & \second{0.3442} & \best{19.1080} & 0.5035 & \best{0.2922} & \best{16.8896} & 0.4645 & \second{0.3026} & \second{14.7497} & 0.4699 & \best{0.3215} & 14.6440 & 0.6092 & \second{0.2702} & \best{18.3918} & 0.4955 & \best{0.2771} \\
\bottomrule
\end{tabular}
\end{adjustbox}
\endgroup
\end{minipage}
\end{table}

\begin{figure*}
    \centering
    \includegraphics[width=0.9\textwidth, keepaspectratio]{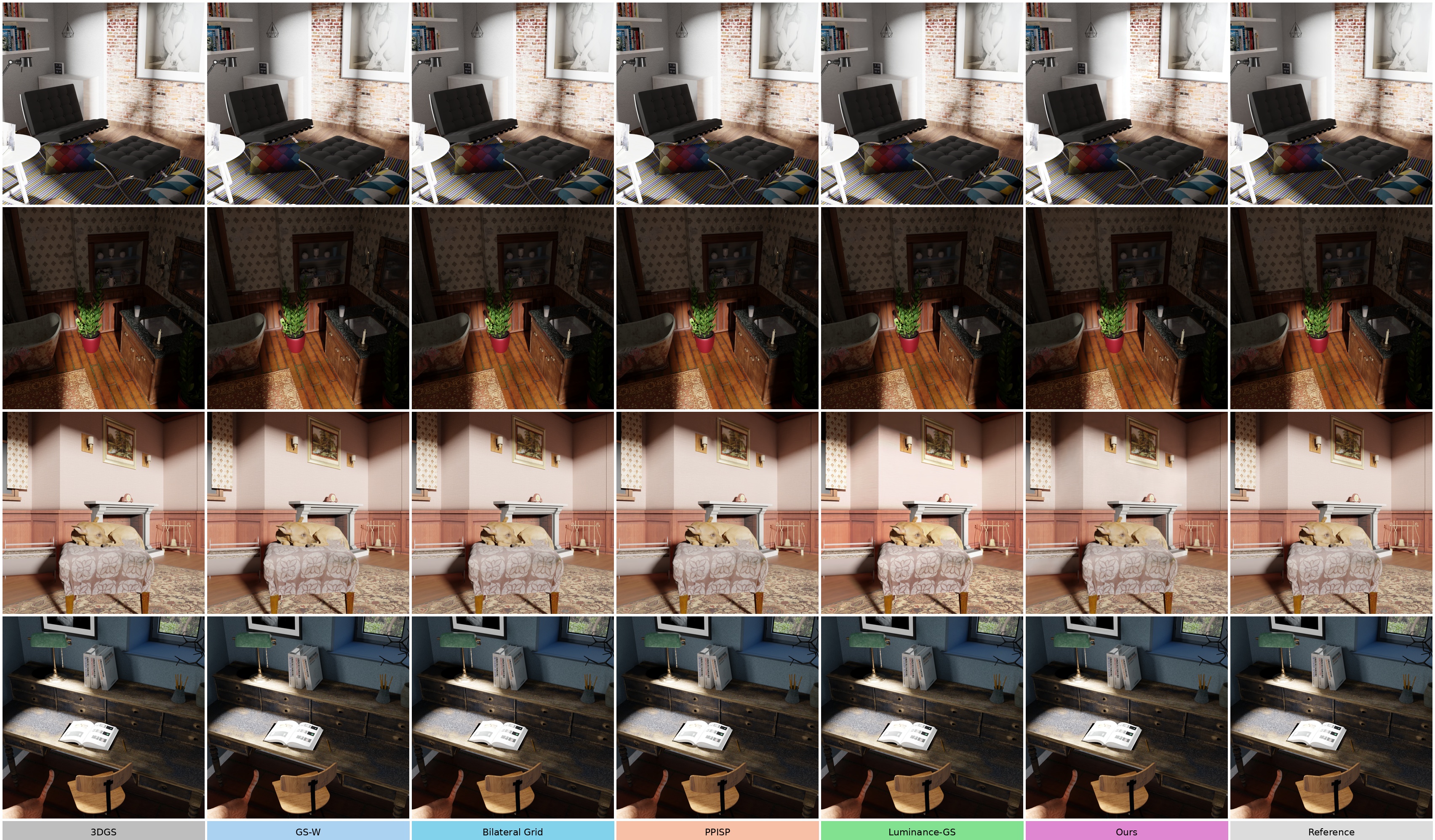}
    \caption{Qualitative comparison of \textit{Rendered Observation} on the
    four HDR-NeRF scenes. Columns show 3DGS, GS-W, Bilateral Grid, PPISP,
    Luminance-GS, EvenSplat, and the reference. The reference retains the
    severe but view-consistent illumination used to generate the observations;
    this comparison therefore measures complete-scene reconstruction fidelity
    rather than illumination removal.}
    \label{fig:hdr-rendered-observation}
\end{figure*}

\begin{figure*}
    \centering
    \includegraphics[width=0.9\textwidth, keepaspectratio]{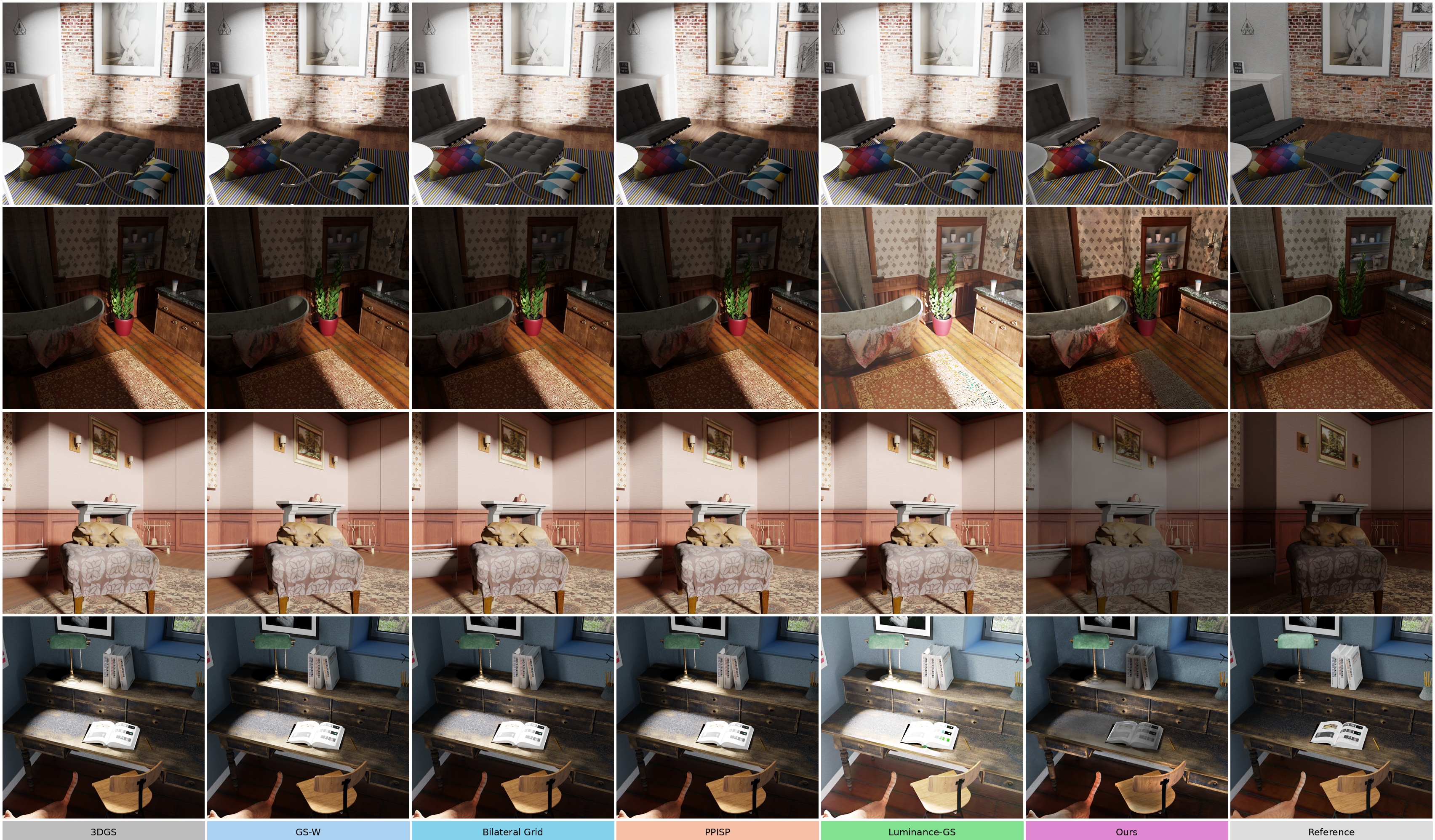}
    \caption{Qualitative comparison of \textit{Base Appearance} recovery on
    the four HDR-NeRF scenes. Columns show 3DGS, GS-W, Bilateral Grid, PPISP,
    Luminance-GS, EvenSplat, and the reference rendered under parallel uniform
    light. EvenSplat directly renders its decomposed base appearance, Luminance-GS produces enhanced renderings, whereas 
    the other comparison methods contribute their standard scene renderings because
    they do not expose a separate base-appearance output.}
    \label{fig:hdr-base-appearance}
\end{figure*}

\paragraph{Additional Experiment on HDR-NeRF dataset.}
\label{app:hdr_nerf}
We evaluate the public simulated HDR-NeRF scenes~\citep{huang2022hdr} under
severe but view-consistent illumination. The \textit{Rendered Observation}
target retains this illumination and measures complete-scene reconstruction (Fig.~\ref{fig:hdr-rendered-observation}).
The \textit{Base Appearance} target instead uses the ground-truth appearance rendered
under parallel uniform light. EvenSplat directly produces this second output; Luminance-GS produces enhanced renderings;
the other comparison methods do not explicitly decompose base appearance, so their
standard renderings are compared with the same target (Fig.~\ref{fig:hdr-base-appearance}).

EvenSplat maintains high-quality full-scene observation rendering, achieving 37.6-dB PSNR on average and remaining visually comparable to methods dedicated
to fitting the illuminated observations. As shown in Table~\ref{tab:hdr-base-appearance}, its additional decomposition output
has the highest base-appearance PSNR on all four scenes, improving mean PSNR
from 13.23~dB for the strongest baseline to 17.91~dB. SSIM and LPIPS remain
mixed, particularly on \textit{CatRoom}, so the improvement should be read as
better recovery of the uniform-light target rather than a universal perceptual
advantage. This is because in regions saturated by the same strong global light in every view,
clamping removes the original color information from the entire observation
set. No reconstruction method can identify those colors without additional
measurements or priors; this limitation comes from the dataset observations,
not from the EvenSplat decomposition.


\begin{table*}[thb]
\centering
\caption{Quantitative recovery of base appearance on HDR-NeRF. Outputs are
evaluated against ground-truth renderings under parallel uniform illumination.
EvenSplat uses its explicit base-appearance output; Luminance-GS uses its enhanced output while other comparison methods use
their standard renderings because they do not separately model this output.}
\label{tab:hdr-base-appearance}
\fontsize{6}{7}\selectfont
\setlength{\tabcolsep}{1.2pt}
\renewcommand{\arraystretch}{1.12}
\begin{adjustbox}{max width=\linewidth}
\begin{tabular}{l*{12}{r}}
\toprule
\textbf{Method}
& \multicolumn{3}{c}{\textbf{CatRoom/desk}}
& \multicolumn{3}{c}{\textbf{BathRoom/bathroom}}
& \multicolumn{3}{c}{\textbf{DogRoom/dog}}
& \multicolumn{3}{c}{\textbf{Sofa/sofa}} \\
\cmidrule(lr){2-4}
\cmidrule(lr){5-7}
\cmidrule(lr){8-10}
\cmidrule(lr){11-13}
& PSNR$\uparrow$ & SSIM$\uparrow$ & LPIPS$\downarrow$
& PSNR$\uparrow$ & SSIM$\uparrow$ & LPIPS$\downarrow$
& PSNR$\uparrow$ & SSIM$\uparrow$ & LPIPS$\downarrow$
& PSNR$\uparrow$ & SSIM$\uparrow$ & LPIPS$\downarrow$ \\
\midrule
3DGS
& 16.1779 & 0.8943 & 0.0989
& 14.7923 & 0.5109 & 0.3258
& 11.1984 & 0.7411 & 0.2668
& 10.7005 & 0.5492 & 0.3190 \\

GS-W
& 16.1751 & 0.8945 & 0.0986
& 14.7767 & 0.5151 & 0.3175
& 11.1819 & 0.7386 & 0.2752
& 10.7238 & 0.5507 & 0.3164 \\

Bilateral Grid
& 16.1449 & 0.8735 & 0.0989
& 14.7900 & 0.5110 & 0.3241
& 11.1873 & 0.7383 & 0.2724
& 10.7103 & 0.5497 & 0.3174 \\

PPISP
& 16.1873 & 0.8992 & 0.0936
& 14.7960 & 0.5185 & 0.3078
& 11.2013 & 0.7489 & 0.2409
& 10.7288 & 0.5516 & 0.3047 \\

Luminance-GS
& 8.9776 & 0.5044 & 0.2877
& 8.3358 & 0.4899 & 0.3275
& 10.3754 & 0.7338 & 0.2808
& 8.4031 & 0.4901 & 0.3164 \\

EvenSplat (Ours)
& 18.3390 & 0.8243 & 0.1550
& 18.3685 & 0.6811 & 0.2863
& 14.4494 & 0.7540 & 0.2483
& 20.4864 & 0.7586 & 0.2346 \\
\bottomrule
\end{tabular}
\end{adjustbox}
\end{table*}

\end{document}

%% file: math_commands.tex
\usepackage{amsmath,amsfonts,bm}

\def\1{\bm{1}}

\DeclareMathAlphabet{\mathsfit}{\encodingdefault}{\sfdefault}{m}{sl}
\SetMathAlphabet{\mathsfit}{bold}{\encodingdefault}{\sfdefault}{bx}{n}

%% file: appendix_losses.tex
\subsection{Method and Loss Details}
\label{app:loss_definitions}

This section details Gaussian compositing, image-space decomposition, and the regularizers grouped in
$\mathcal{R}$ in Eq.~\eqref{eq:total_objective}. 
% Observation fidelity and
% cross-domain coupling are defined in Sec.~\ref{subsec:coupling}; the grouped
% objective counts observation reconstruction once.

\subsubsection{Gaussian Compositing and Appearance Estimates}
\label{app:rendering_details}

For the ordered Gaussians contributing to pixel $\mathbf{p}$, the standard
front-to-back compositing weight is
\begin{equation}
w_j(\mathbf{p})=T_j(\mathbf{p})\alpha_j(\mathbf{p}),
\qquad
T_j(\mathbf{p})=\prod_{k<j}(1-\alpha_k(\mathbf{p})).
\label{eq:compositing_weight}
\end{equation}
Here $\alpha_j(\mathbf{p})$ is Gaussian $j$'s opacity contribution at the
pixel and $T_j(\mathbf{p})$ is the transmittance through preceding Gaussians.
The same weights render illumination-modulated color and illumination in
Eq.~\eqref{eq:rendered_illumination_map}. 
% If the optional additive-residual
% branch is enabled, the field outputs
% $(\mathbf{a}_j,\mathbf{b}_j)=\mathcal{F}_{\mathrm{illum}}
% (\boldsymbol{\mu}_j,\mathbf{n}_j,\mathbf{v}_j)$ and the color rendering becomes
% \begin{equation}
% \hat{\mathbf{I}}_{\mathrm{lit}}(\mathbf{p})
% =\sum_j w_j(\mathbf{p})
% \left(\mathbf{a}_j\odot\mathbf{C}_j^r+\mathbf{b}_j\right).
% \label{eq:rendered_lit_image_with_bias}
% \end{equation}
% Only the multiplicative coefficients $\mathbf{a}_j$ enter
% $\mathbf{L}_{3D}$. The additive residual remains in the rendered image
% after division by this multiplicative illumination map.

The base-appearance quotient documented in the preceding formulation is
\begin{equation}
\hat{\mathbf{R}}_{\mathrm{3D}}
=\hat{\mathbf{I}}_{\mathrm{adapt}}
\oslash(\mathbf{L}_{adapt}+\epsilon).
\label{eq:reflectance_division}
\end{equation}
A quotient of composited maps generally differs
from $\sum_j w_j\mathbf{C}_j^r$ during training, it also retains contributions from image-specific corrections.
% These contributions limit the quotient's interpretation to an appearance
% estimate during training rather than calibrated reflectance. 
% \CW{[The denominator and its relationship to local
% compensation remain subject to the implementation checks in
% Appendix~\ref{app:method_verification}]}.

\subsubsection{Image-Space Architecture and Losses}
\label{app:image_losses}

% \CW{[TODO: Fix the image-space network figure \ref{fig:short}, many modules are missing in this figure. Take Jacob's version as refernce.]}
\paragraph{Image-Space Network.}
Given an input image $\mathbf{I}^{i}$ and a view-specific embedding $\mathbf{e}^{i}$, the image-space network uses a lightweight U-Net-style~\citep{ronneberger2015u} architecture with Spatially-FiLM~\citep{perez2017filmvisualreasoninggeneral} conditioning, Retinex blocks, and channel--spatial attention. Multi-scale encoder features are fused with decoder features through skip connections to predict the illumination map $\mathbf{L}_{2D}^{i}$. The base appearance is then obtained by the element-wise division in Eq.~\eqref{eq:retinex2d_reflectance}.

\begin{figure*}[htb]
    \centering
    \includegraphics[width=0.90\linewidth]{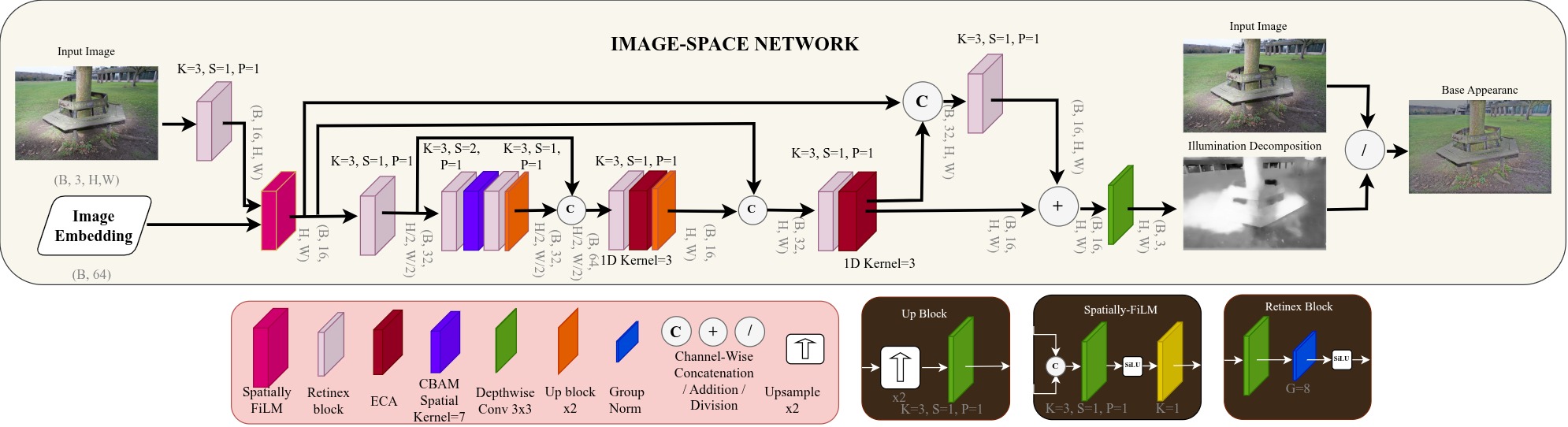}
    \caption{Architecture of the image-space network for base-appearance and illumination decomposition.}
    \label{fig:short}
\end{figure*}

The image-space branch predicts illumination $\mathbf{L}_{2D}$ and derives
base appearance $\mathbf{R}_{2D}$. Its intrinsic regularization groups the
three perspectives illustrated in Fig.~\ref{fig:decomposition_priors}:
\begin{equation}
\mathcal{L}_{\mathrm{prior}}
=\lambda_{\mathrm{curve}}\mathcal{L}_{\mathrm{curve}}
+\lambda_{\mathrm{smooth}}\mathcal{L}_{\mathrm{smooth}}^{2D}
+\lambda_{\mathrm{white}}\mathcal{L}_{\mathrm{white}}.
\label{eq:image_prior_group}
\end{equation}
% The $\omega$ coefficients denote the corresponding prior weights.
Eq.~\eqref{eq:coupling_objective} connects this branch to the scene through
illumination alignment and image recombination. Below, $N$ denotes the number
of pixels, and $\nabla_x$ and $\nabla_y$ denote horizontal and vertical image
gradients. Scalar intensity targets are broadcast to the color channels; if a
selected set is empty, its corresponding term is taken as zero.

\paragraph{Adaptive Curve Loss.}
The adaptive curve loss constrains overly dark and bright base-appearance
regions:
\begin{equation}
\begin{aligned}
\mathcal{L}_{\mathrm{curve}}
=& \eta_{\mathrm{dark}}\frac{1}{|\Omega_{\mathrm{dark}}|}
\sum_{p\in\Omega_{\mathrm{dark}}}|\mathbf{R}_{2D}(p)-\rho_{\mathrm{dark}}| \\
&+ \eta_{\mathrm{bright}}\frac{1}{|\Omega_{\mathrm{bright}}|}
\sum_{p\in\Omega_{\mathrm{bright}}}|\mathbf{R}_{2D}(p)-\rho_{\mathrm{bright}}| 
% &+ \eta_{\mathrm{grad}}\frac{1}{N}\sum_p
% \left(|\nabla_x\mathbf{R}_{2D}(p)|^2+|\nabla_y\mathbf{R}_{2D}(p)|^2\right).
\end{aligned}
\label{eq:loss_adaptive_curve}
\end{equation}
Here $\Omega_{\mathrm{dark}}$ and $\Omega_{\mathrm{bright}}$ select
base-appearance intensities below $T_{\mathrm{dark}}$ and above
$T_{\mathrm{bright}}$, respectively. The constants $\rho_{\mathrm{dark}}$
and $\rho_{\mathrm{bright}}$ are the preferred intensities for the two sets:
the thresholds select where the loss applies, while the targets determine
the direction of correction. The coefficients $\eta_{\mathrm{dark}}$ and
$\eta_{\mathrm{bright}}$ weight the two target
penalties within this loss. The single-branch settings
in Table~\ref{tab:retinex2d-hyperparams} use targets 0.45 and 0.55 with
selection thresholds 0.25 and 0.65. These target symbols leave $\alpha_j$
reserved for Gaussian opacity.

\paragraph{Edge-Aware Smoothness Loss.}
To encourage smooth illumination while preserving image edges, we use
\begin{equation}
\mathcal{L}_{\mathrm{smooth}}^{2D}
= \frac{1}{N}\sum_{p,d}
\exp\left(-\frac{|\nabla_d\mathbf{I}_{\mathrm{gray}}(p)|}{\gamma_{\mathrm{edge}}}\right)
|\nabla_d\mathbf{L}_{2D,\mathrm{gray}}(p)|,
\label{eq:loss_edge_aware_smooth}
\end{equation}
Here $d\in\{x,y\}$ indexes the gradient direction, and the subscript
$\mathrm{gray}$ denotes a grayscale map. The parameter
$\gamma_{\mathrm{edge}}>0$ sets the input-gradient scale at which the
smoothness weight decays. A smaller value reduces smoothing across weaker
image edges, while a larger value retains smoothing across stronger edges.

\paragraph{White Preservation Loss.}
A soft mask identifies bright and approximately achromatic regions in the
input image's CIELAB representation:
\begin{equation}
M_{\mathrm{white}}(p)
= \sigma\left(\kappa_{\mathrm{white}}(L^{*}(p)-T_{\mathrm{white}})\right)
\exp\left(-\frac{(a^{*}(p))^2+(b^{*}(p))^2}{2s_{\mathrm{chroma}}^2}\right).
\label{eq:white_mask}
\end{equation}
Here $L^{*}$ is CIELAB lightness, and $a^{*}$ and $b^{*}$ are its two
opponent-color coordinates; values near zero indicate low chroma.
$T_{\mathrm{white}}$ is the lightness threshold,
$\kappa_{\mathrm{white}}>0$ controls the sigmoid transition steepness, and
$s_{\mathrm{chroma}}>0$ is the chroma tolerance. Increasing
$s_{\mathrm{chroma}}$ admits more chromatic pixels into the soft mask.
These parameters operate in the chosen CIELAB coordinate scale; the
brightness targets $\rho_{\mathrm{dark}}$ and $\rho_{\mathrm{bright}}$
operate in the base-appearance image domain.
The corresponding loss is
\begin{equation}
\mathcal{L}_{\mathrm{white}}
= \frac{1}{N}\sum_p M_{\mathrm{white}}(p)
\left|\mathbf{R}_{2D}(p)-\mathbf{I}(p)\right|.
\label{eq:loss_white_preservation}
\end{equation}
The mask expresses a preference for preserving bright, approximately
achromatic input regions. Clipped highlights may also receive high mask values,
which limits the reliability of this heuristic in saturated regions.

\subsubsection{Gaussian-Level Losses}
\label{app:gaussian_losses}

The Gaussian-level branch uses reconstruction supervision and regularization
of appearance and illumination, together with the cross-domain coupling
described in the main text.

\paragraph{Observation Reconstruction.}
The illumination-aware rendering uses the observation loss from the main text:
\begin{equation}
\mathcal{L}_{\mathrm{obs}}
=D(\hat{\mathbf{I}}_{\mathrm{adapt}},\mathbf{I}).
\label{eq:loss_3d_reconstruction_gaussian}
\end{equation}
It has already been counted in Eq.~\eqref{eq:total_objective}.

% \paragraph{Exclusion Loss.}
% To encourage separation of base appearance and illumination, we penalize
% correlated gradients between their rendered maps:
% \begin{equation}
% \mathcal{L}_{\mathrm{excl}}
% = \mathcal{E}\left(\hat{\mathbf{R}}_{\mathrm{3D}},\mathbf{L}_{\mathrm{adapt}}\right),
% \label{eq:loss_exclusion}
% \end{equation}
% where $\mathcal{E}(\cdot)$ denotes the gradient-exclusion operator.

\paragraph{Exclusion Loss.}
To encourage separation of base appearance and illumination, we penalize
correlated gradients between their rendered maps:
\begin{equation}
\mathcal{L}_{\mathrm{excl}}
= \mathcal{E}\left(\hat{\mathbf{R}}_{\mathrm{3D}},\mathbf{L}_{\mathrm{adapt}}\right),
\label{eq:loss_exclusion}
\end{equation}
where the gradient-exclusion operator follows the exclusion loss of
\citet{zhang2018single}, defined over a multi-scale pyramid as
\begin{equation}
\mathcal{E}(\mathbf{R},\mathbf{L})
=
\sum_{l=0}^{L-1}
\sum_{d\in\{x,y\}}
\frac{1}{N_l}
\sum_{p}
\left|
\phi\!\left(\nabla_d \mathbf{R}^{\,l}(p)\right)
\cdot
\phi\!\left(\nabla_d \mathbf{L}^{\,l}(p)\right)
\right|,
\label{eq:gradient_exclusion}
\end{equation}
where $\mathbf{R}^{\,l}$ and $\mathbf{L}^{\,l}$ denote the reflectance and
illumination maps at pyramid level $l$, respectively, $N_l$ is the number of
pixels at that level, $\nabla_d$ denotes the spatial gradient along
$d\in\{x,y\}$, and $\phi(z)=\tanh(z)$ suppresses small gradient noise and
normalizes the gradient responses.

\paragraph{Higher-Order SH Regularization.}
We regularize higher-order spherical harmonic coefficients to discourage
illumination variations from being absorbed into Gaussian appearance:
\begin{equation}
\mathcal{L}_{\mathrm{SH}}
= \frac{1}{N_{\mathrm{SH}}}\sum_{i=1}^{N_{\mathrm{SH}}}c_i^2,
\label{eq:loss_sh_regularization}
\end{equation}
where $c_i$ denotes a higher-order SH coefficient.

\subsubsection{ILEC Regularization}
\label{app:local_losses}

\paragraph{Relative Illumination Guidance.}
Let $\hat{\Delta}_{\mathrm{loc}}^{i}
=\mathrm{Up}(\Delta_{\mathrm{loc}}^{i})$ denote the bilinearly upsampled
log-compensation map. For each color channel $c$, we normalize the rendered
illumination over the current image:
\begin{equation}
\mathbf{L}_{\mathrm{rel}}^{c}
=\frac{\mathbf{L}_{3D}^{c}
-\min_{\mathbf{p}}\mathbf{L}_{3D}^{c}(\mathbf{p})}
{\max_{\mathbf{p}}\mathbf{L}_{3D}^{c}(\mathbf{p})
-\min_{\mathbf{p}}\mathbf{L}_{3D}^{c}(\mathbf{p})+\epsilon}.
\label{eq:relative_illumination}
\end{equation}
The guidance map is
\begin{equation}
\mathbf{M}_{\mathrm{rel}}
=m_{\mathrm{floor}}+(1-m_{\mathrm{floor}})
\sigma\left(\kappa_{\mathrm{guide}}
(\mathbf{L}_{\mathrm{rel}}-t_{\mathrm{guide}})\right),
\label{eq:relative_guidance}
\end{equation}
Here $m_{\mathrm{floor}}\in[0,1)$ is the minimum guidance weight,
$\kappa_{\mathrm{guide}}>0$ controls transition steepness, and
$t_{\mathrm{guide}}\in[0,1]$ is the threshold on normalized illumination.
Larger values of $\mathbf{M}_{\mathrm{rel}}$ identify regions that are
relatively bright within the image; smaller values identify regions receiving
less guidance. The map weights the residual regularizers below, and
Eq.~\eqref{eq:adapted_illumination} defines its forward exposure adaptation.
Image indices are omitted for readability. Products and absolute values in
the following expressions are evaluated element-wise, with channel reduction
implicit in the image averages.

\paragraph{Residual Magnitude Penalty.}
An $\ell_1$ penalty limits the overall amplitude of the log-exposure residual
and favors zero correction where the other objectives provide little evidence
for an adjustment:
\begin{equation}
\mathcal{L}_{\mathrm{mag}}
= \frac{1}{N}\sum_p\left|\hat{\Delta}_{\mathrm{loc}}^{i}(p)\right|.
\label{eq:loss_residual_magnitude}
\end{equation}

\paragraph{Guidance-Weighted Shrinkage.}
This term adds stronger shrinkage where $\mathbf{M}_{\mathrm{rel}}$ is small:
\begin{equation}
\mathcal{L}_{\mathrm{shrink}}
= \frac{1}{N}\sum_p
\left|\left(1-\mathbf{M}_{\mathrm{rel}}(p)\right)
\hat{\Delta}_{\mathrm{loc}}^{i}(p)\right|.
\label{eq:loss_guided_shrinkage}
\end{equation}
The continuous weight $1-\mathbf{M}_{\mathrm{rel}}$ concentrates correction
in the more strongly guided regions. It provides a soft spatial preference
whose strength varies across the image.

\paragraph{Bright-Region Exposure Promotion.}
This signed term favors positive log-exposure residuals where
$\mathbf{M}_{\mathrm{rel}}$ is large:
\begin{equation}
\mathcal{L}_{\mathrm{promote}}
= -\frac{1}{N}\sum_p\mathbf{M}_{\mathrm{rel}}(p)
\hat{\Delta}_{\mathrm{loc}}^{i}(p).
\label{eq:loss_exposure_promotion}
\end{equation}
For a fixed guidance map, a positive residual decreases this term and
increases the corresponding multiplicative factor in
Eq.~\eqref{eq:adapted_illumination}. It encourages the exposure-adaptation
branch to account for additional illumination in relatively bright regions.
Its effect is balanced by the magnitude and shrinkage penalties and the
remaining training objective.

The combined local regularization objective is
\begin{equation}
\mathcal{L}_{\mathrm{loc}}
= \lambda_{\mathrm{mag}}\mathcal{L}_{\mathrm{mag}}
+ \lambda_{\mathrm{shrink}}\mathcal{L}_{\mathrm{shrink}}
+ \lambda_{\mathrm{promote}}\mathcal{L}_{\mathrm{promote}}.
\label{eq:local_regularization}
\end{equation}
The local weights are distinguished from the SSIM mixture coefficient
$\lambda_{\mathrm{ssim}}$ in Eq.~\eqref{eq:image_discrepancy}.